%% file: main.tex
\documentclass[runningheads]{llncs}

\usepackage{eccv}

\usepackage{eccvabbrv}

\usepackage{graphicx}
\usepackage{booktabs}
\usepackage{array}
\usepackage{adjustbox}

\usepackage{xcolor}         
\usepackage{graphicx}
\usepackage{wrapfig}
\usepackage{mathtools}
\usepackage{enumitem}
\usepackage{subcaption}
\usepackage{algorithm}
\usepackage{algpseudocode}
\usepackage{multirow}
\usepackage{soul}    
\usepackage{todonotes}
\usepackage{kotex}

\usepackage[accsupp]{axessibility}  

\usepackage{hyperref}

\usepackage{orcidlink}

\makeatletter
\def\blfootnote{\gdef\@thefnmark{}\@footnotetext}
\makeatother

\begin{document}

\title{Unlearning for One-Step Generative Models \\
via Unbalanced Optimal Transport} 

\titlerunning{Unlearning for One-Step Generative Models via UOT}

\author{Hyundo Choi\inst{1\star}\orcidlink{0009-0009-6452-1204} \and
Junhyeong An\inst{1\star}\orcidlink{0009-0008-1970-7459} \and
Jinseong Park\inst{2\dagger}\orcidlink{0000-0003-1931-8441} \and
Jaewoong Choi\inst{1\dagger}\orcidlink{0009-0004-1980-011X}}

\blfootnote{$^\star$ Equal contribution.}
\blfootnote{$^\dagger$ Corresponding author.}

{
  \let\thefootnote\relax
  \footnotetext{Preprint. Under review.}
}

\authorrunning{H. Choi et al.}

\institute{Sungkyunkwan University, Seoul, Republic of Korea \and
Korea Institute for Advanced Study, Seoul, Republic of Korea \\
\email{\{hyuncde12, jhan0429, jaewoongchoi\}@skku.edu, jinseong@kias.re.kr}}

\maketitle

\begin{abstract}
Recent advances in one-step generative frameworks, such as flow map models, have significantly improved the efficiency of image generation by learning direct noise-to-data mappings in a single forward pass. However, machine unlearning for ensuring the safety of these powerful generators remains entirely unexplored. Existing diffusion unlearning methods are inherently incompatible with these one-step models, as they rely on a multi-step iterative denoising process. In this work, we propose UOT-Unlearn, a novel plug-and-play class unlearning framework for one-step generative models based on the Unbalanced Optimal Transport (UOT). Our method formulates unlearning as a principled trade-off between a forget cost, which suppresses the target class, and an $f$-divergence penalty, which preserves overall generation fidelity via relaxed marginal constraints. By leveraging UOT, our method enables the probability mass of the forgotten class to be smoothly redistributed to the remaining classes, rather than collapsing into low-quality or noise-like samples.  Experimental results on CIFAR-10 and ImageNet-256 demonstrate that our framework achieves superior unlearning success (PUL) and retention quality (u-FID), significantly outperforming baselines.
\end{abstract}

\input{text}


\section*{Acknowledgements}
Hyundo Choi, Junhyeong An, and Jaewoong Choi are partially supported by the National Research Foundation of Korea (NRF) grant funded by the Korea government (MSIT) (No. RS-2024-00349646). Jinseong Park is supported by a KIAS Individual Grant (AP102301, AP102303) via the Center for AI and Natural Sciences at Korea Institute for Advanced Study. This work was supported by the Center for Advanced Computation at Korea Institute for Advanced Study.

%
%
\bibliographystyle{splncs04}
\bibliography{main}

\clearpage
\appendix
\input{text_sup}

\end{document}

%% file: text.tex
\section{Introduction}

Generative models, particularly diffusion models, have achieved high-quality image synthesis \cite{ho2020denoising, song2020score, song2021denoising}. However, their practical utility is heavily bottlenecked by slow inference speeds, which stem from the requirement of tens to hundreds of iterative denoising steps. To overcome this limitation, recent advances have rapidly shifted towards one-step generative architectures, such as consistency models or flow maps \cite{lipman2023flow, liu2022flow, song2023consistency, albergo2023building}. By directly mapping the noise distribution to the data distribution in a single forward pass, these models achieve near-diffusion-level generation quality with an advantage in sampling speed.

As these generative models grow faster and more powerful, the risk of producing undesirable content, such as Not Safe For Work (NSFW) imagery or copyrighted materials, has simultaneously amplified. To mitigate these risks without the prohibitive cost of retraining models from scratch, machine unlearning has emerged as an essential safeguard \cite{bourtoule2021machine, Nguyen2025Unlearning}. While unlearning techniques have been actively developed for standard multi-step diffusion models, the field of one-step generative models remains entirely unexplored. This gap is particularly concerning because the extreme generation speed of one-step models can drastically accelerate the spread of harmful content. Therefore, establishing an unlearning framework for one-step generators needs to be investigated.


Crucially, existing diffusion unlearning methods \cite{kumari2023ablating, zhang2024forget, fan2024salun, gandikota2023erasing} cannot be straightforwardly applied to one-step architectures. Previous techniques inherently rely on multi-step denoising processes, tweaking noise predictions or gradients at specific intermediate timesteps. In contrast, one-step models map noise to data in a single forward pass. Without intermediate steps during sampling, traditional step-by-step modifications are difficult to apply in one-step generators \cite{geng2025mean}.

To bridge this critical gap, we propose \textbf{\textit{UOT-Unlearn}}, the first plug-and-play class unlearning framework designed specifically for one-step generative models. Our framework addresses the unlearning problem by exploiting the Unbalanced Optimal Transport (UOT). 
Unlike the standard Optimal Transport (OT), which enforces strict distribution matching, UOT relaxes the marginal constraints and instead minimizes a trade-off between the transport cost and distributional deviation. This flexibility enables us to control the balance between removing the forget class and preserving the overall data distribution. In particular, we introduce an unlearning cost that penalizes samples belonging to the forget concept. This heavy penalization induces distribution mismatch for the forget concept, which leads to the unlearning. Based on the neural optimal transport formulation for UOT, we derive the unlearning objective that fine-tunes one-step generative models. Importantly, UOT-Unlearn operates using only generated samples and a forget centroid, eliminating the need for real retain datasets while preserving the overall generation quality.
\begin{itemize}
    \item We introduce UOT-Unlearn, the first unlearning framework tailored for one-step generative models based on the optimal transport formulation. 
    \item We formulate a novel UOT-based objective that smoothly redistributes the target class probability into the remaining classes via an $f$-divergence penalty.
    \item Experiments on benchmark datasets (e.g., CIFAR-10, ImageNet-256) using representative one-step architectures, such as CTM and Meanflow models, demonstrate that our method achieves superior class unlearning (PUL) and retention quality (u-FID).
\end{itemize}

\section{Preliminaries}

\subsection{One-Step Generative Models via Probability Flow}
Continuous-time generative models, such as diffusion models \cite{song2020score} and flow matching \cite{lipman2023flow, liu2022flow}, aim to learn a continuous transformation between a tractable noise distribution $p_{0}$ and a target data distribution $p_{1}=p_{\mathrm{data}}$. This transformation is represented by the Ordinary Differential Equation (ODE) modeling the probability flow:
\begin{equation} \label{eq:pf_ode}
\mathrm{d}x_t = v_\theta(x_t, t) \mathrm{d}t, \quad t \in [0, 1],
\end{equation}
where $v_\theta$ denotes a time-dependent velocity field. In diffusion models, $v_\theta$ is implicitly defined via the score function (PF-ODE) \cite{song2020score}. In flow matching, it is learned through a direct regression to the conditional vector field. Generating samples from these models requires numerically integrating this ODE from $t=0$ to $t=1$. This iterative process requires tens to hundreds of neural network evaluations, making inference computationally expensive.

To overcome this limitation, recent advances in Flow Map generative modeling \cite{geng2025mean, song2023consistency,kim2023consistency} aim to directly learn the solution to the probability flow (\cref{eq:pf_ode}), thereby enabling one-step or few-step generation. Formally, models like Consistency Trajectory Model (CTM) \cite{kim2023consistency} and MeanFlow \cite{geng2025mean} distill or explicitly parameterize the flow map $\psi(x_t, t, s)$, which maps a state at time $t$ directly to time $s$:
\begin{equation}
    \psi(x_t, t, s) = x_t + \int_{t}^{s} v_\theta(x_\tau, \tau) \mathrm{d}\tau.
\end{equation}
By learning this mapping directly, the entire generation process can be executed in a single forward pass: 
\begin{equation}
    x_1 = \psi(x_0, 0, 1) = G_\theta(x_0).
\end{equation}
Existing machine unlearning methods for generative models are largely designed for multi-step diffusion processes \cite{kumari2023ablating, zhang2024forget, fan2024salun, gandikota2023erasing}, where modifications are applied at intermediate denoising steps. Such approaches are inherently incompatible with one-step generative architectures. UOT-Unlearn addresses this gap by intervening strictly at the final mapping stage $G_\theta(x_0)$. This trajectory-agnostic approach allows our method to be seamlessly integrated into any pretrained one-step model.



\subsection{Unbalanced Optimal Transport (UOT)}
Optimal Transport (OT) provides a principled framework for finding a cost-efficient mapping that transforms a source distribution $\mu$ into a target distribution $\nu$ \cite{villani2009optimal, santambrogio2015optimal}.
In the Kantorovich formulation \cite{kantorovich1948monge}, this is defined by searching for a joint probabilistic coupling $\pi$ that bridges these two distributions while minimizing the total transport cost \cite{villani2009optimal}:
\begin{equation}
C(\mu,\nu) := \inf_{\pi\in\mathrm{\Pi}(\mu,\nu)} \left[ \int_{\mathcal{X}\times\mathcal{Y}} c(x,y) \, \mathrm{d}\pi(x,y) \right],
\label{eq:ot}
\end{equation}
where $\mathrm{\Pi}(\mu,\nu)$ denotes the set of all joint probability distributions whose marginals $\pi_0$ and $\pi_1$ must \textbf{exactly match} the source $\mu$ and target $\nu$, respectively \cite{villani2009optimal}. 
Although OT provides a mathematically principled framework for distribution alignment, the strict marginal constraints can make the resulting transport plan overly rigid. In scenarios where probability mass must be removed or redistributed, such as machine unlearning, this rigidity becomes problematic.

Unbalanced Optimal Transport (UOT) addresses this limitation by relaxing the hard marginal constraints through divergence penalties \cite{uot1, uot2}.
Instead of enforcing exact marginal matching, UOT minimizes a \textbf{principled trade-off} between the transport cost and the marginal deviation \cite{balaji2020robust, choi2023generative}:
\begin{equation}
C_{ub}(\mu,\nu) := \inf_{\pi\in\mathcal{M}_{+}(\mathcal{X}\times\mathcal{Y})} \left[ \int_{\mathcal{X}\times\mathcal{Y}} c(x,y) \, \mathrm{d}\pi(x,y) + D_{\mathrm{\Psi}_1}(\pi_0|\mu) + D_{\mathrm{\Psi}_2}(\pi_1|\nu) \right],
\label{eq:uot}
\end{equation}
where $\mathcal{M}_{+}(\mathcal{X}\times\mathcal{Y})$ denotes the set of positive measures defined on $\mathcal{X} \times \mathcal{Y}$, and $\pi_0, \pi_1$ are the marginal distributions of $\pi$. 
The two $f$-divergences $D_{\mathrm{\Psi}}$ measure the discrepancy between the marginals ($\pi_{i}$) and the corresponding source and target distributions ($\mu$ and $\nu$).
Formally, for a convex, lower semi-continuous, and non-negative entropy function $\mathrm{\Psi}$, the $f$-divergence between the marginal $\pi_0$ and the source distribution $\mu$ is defined as:
\begin{equation}
D_{\mathrm{\Psi}_1}(\pi_0|\mu) = \int_{\mathcal{X}} \mathrm{\Psi}_1\left(\frac{\mathrm{d}\pi_0(x)}{\mathrm{d}\mu(x)}\right) \, \mathrm{d}\mu(x),
\label{eq:csiszar}
\end{equation}
and similarly for $D_{\mathrm{\Psi}_2}(\pi_1|\nu)$. 
Notably, UOT serves as a generalization of OT; if $\mathrm{\Psi}_1$ and $\mathrm{\Psi}_2$ are chosen as convex indicator functions of $\{1\}$, the formulation precisely recovers the classical OT problem as any marginal mismatch results in infinite cost \cite{choi2023generative}.

Choi \etal~\cite{choi2023generative} proposed a neural optimal transport algorithm, called UOTM, for learning the optimal transport map of the UOT problem. By leveraging the semi-dual form of UOT, UOTM introduces the following learning objective where the transport map $T_\theta$ and the dual potential $v_\phi$ are parameterized by neural networks:
\begin{equation}
\inf_{v_\phi} \left[ \int_{\mathcal{X}} \mathrm{\Psi}_1^* \left( - \inf_{T_\theta} [c(x, T_\theta(x)) - v_\phi(T_\theta(x))] \right) \mathrm{d}\mu(x) + \int_{\mathcal{Y}} \mathrm{\Psi}_2^* (-v_\phi(y)) \, \mathrm{d}\nu(y) \right],
\label{eq:uot_neural}
\end{equation}
where $\mathrm{\Psi}^*$ denotes the convex conjugate of $\mathrm{\Psi}$. After training, the transport network $T_\theta$ learns an unbalanced optimal transport map between the source distribution $\mu$ and the target distribution $\nu$.


In this work, we leverage the intrinsic trade-off in the UOT formulation between the transport cost and the marginal error to develop a novel unlearning algorithm. 
Intuitively, the UOT objective (\cref{eq:uot}) allows marginal mismatches when the resulting reduction in the transport cost outweighs the corresponding increase in the $f$-divergence penalties. This property is particularly suitable for machine unlearning. By designing a cost function that penalizes the generation of unlearn target, the UOT framework encourages the model to shift probability mass away from the \emph{forget class} while maintaining the overall integrity of the remaining distribution through a relaxed matching process (\cref{sec:unlearn_via_uot}).

\section{Proposed Method} \label{sec:method}
\subsection{Problem Formulation} \label{sec:prob_form}

We consider the problem of \textbf{class unlearning for a one-step generative model}. Let $G_{\mathrm{pre}}: \mathcal{Z} \rightarrow \mathcal{X}$ denote a pretrained one-step generative model that maps a prior noise distribution $x_0 \sim \mu_Z$ to the learned data distribution $p_{\mathrm{pre}} = (G_{\mathrm{pre}})_{\#} \mu_{Z}$. 
The model is trained on the full data distribution $p_{\mathrm{data}}$, which includes both the target concept to be removed (the \textit{forget class}) and all other concepts (the \textit{remaining classes}). Our goal is to fine-tune $G_{\mathrm{pre}}$ into a new generator $G_\theta$ that removes the forget class while preserving the generation quality and diversity of the remaining classes.

Formally, let $\mathcal{S}_f \subset \mathcal{X}$ denote the semantic support in the data space corresponding to the forget class, and let $\mathcal{S}_r \subset \mathcal{X}$ denote the semantic support for the remaining classes.
The goal of our machine unlearning framework is to suppress the probability of generating forget samples while maintaining generation within the retain region. This objective can be expressed as
\begin{equation}
\mathbb{P}\big(G_\theta(x_0) \in \mathcal{S}_f\big) \to 0, \quad \text{and ideally,} \quad \mathbb{P}\big(G_\theta(x_0) \in \mathcal{S}_r\big) \to 1.
\label{eq:unlearn_goal}
\end{equation}
Our framework aims to achieve this objective for one-step generators operating in a single forward pass, without requiring access to real retain data during the unlearning optimization.

\subsection{Unlearning via Unbalanced Optimal Transport} \label{sec:unlearn_via_uot}

We formulate the machine unlearning process as a distribution transportation problem using the Unbalanced Optimal Transport (UOT) framework. 
The key idea is to exploit the intrinsic trade-off between the transportation cost and the distribution matching error.
Specifically, by designing a cost function $c_{\mathrm{ul}}(\cdot, \cdot)$ that imposes a heavy penalty on the \textit{forget} region (detailed in \cref{sec:unlearning_cost}), 
we force the transport plan to avoid generating penalized concepts. 
To mitigate this massive cost, the UOT objective naturally allows for a distribution matching error, safely steering the generative pathway toward an \textit{unlearned} distribution.

To formalize this framework, we consider the following UOT problem where the source is the pretrained distribution, i.e., $\mu = p_{\mathrm{pre}}$, and the target distribution is the full data distribution $\nu = p_{\mathrm{data}}$.
\begin{equation}
\inf_{\pi\in\mathcal{M}_{+}(\mathcal{X}\times\mathcal{Y})} \left[ \int_{\mathcal{X}\times\mathcal{Y}} c(x,y) \, \mathrm{d}\pi(x,y) + D_{\mathrm{\Psi}_1}(\pi_0| p_{\mathrm{pre}}) + D_{\mathrm{\Psi}_2}(\pi_1| p_{\mathrm{data}}) \right],
\label{eq:uot_unlearn}
\end{equation}
In this formulation, the source distribution $\mu$ represents the starting point of the unlearning process. The transported marginal $\pi_{1}$ corresponds to the distribution produced by the updated generator after unlearning. 

The UOT framework induces two desirable properties for $\pi_1$.
First, the divergence term $D_\mathrm{{\Psi_2}}(\pi_1| p_{\mathrm{data}})$ encourages $\pi_1$ to remain close to the data distribution, thereby preserving the overall generation fidelity. Second, when the transport cost assigns a large penalty to the forget region $\mathcal{S}_f$, the optimal transport plan avoids placing probability mass in this region. As a result, the optimal transported marginal $\pi_1^\star$ satisfies $\mathcal{S}_f \not\subset \mathrm{supp}(\pi_1^\star)$,
which effectively suppresses the generation of the forget class and leads to
\begin{equation}
    \mathbb{P}\big(G_\theta(x_0) \in \mathcal{S}_f\big) \rightarrow 0.
\end{equation}

At the same time, the divergence regularization ensures that the redistributed probability mass remains within the support of the data distribution. In particular, when $D_{\mathrm{\Psi_2}}$ is set to the Kullback--Leibler divergence (as used in our experiments), $D_{\mathrm{\Psi}_2}(\pi_1| p_{\mathrm{data}})$ corresponds to a reverse KL divergence, which exhibits the mode seeking property \cite{murphy2022probabilistic}.
This divergence heavily penalizes probability mass assigned to regions where the data distribution has negligible density \cite{murphy2022probabilistic}. In other words, the optimal unlearned distribution $\pi_1^\star$ does not assign positive mass outside $\mathcal{S}_{f} \cup \mathcal{S}_{r}$.
As a result, the redistributed mass concentrates on valid semantic regions corresponding to the remaining classes, leading to
\begin{equation}
\mathbb{P}\big(G_\theta(x_0) \in \mathcal{S}_r\big) \rightarrow 1.
\end{equation}

Based on this formulation, we derive the unlearning objective using the semi-dual UOT framework. Let $\Delta T$ denote the unbalanced optimal transport map that transforms the pretrained distribution $p_{\mathrm{pre}}$ to the unlearned distribution $\pi_1$. Following the UOTM formulation (\cref{eq:uot_neural}), we parameterize this transport map by a neural network $\Delta T_\theta$ and obtain the following learning objective:
\begin{multline} \label{eq:uot_unlearn1}
\inf_{v_\phi}\ \Bigg[ \int_{\mathcal{X}} \mathrm{\mathrm{\Psi}}_1^{*}\!\Big( -\inf_{\Delta T_\theta}\big\{ c(x_1, \Delta T_{\theta}(x_1)) - v_\phi( \Delta T_\theta(x_1)) \big\} \Big) \mathrm{d}p_{\mathrm{pre}}(x_1)\\ 
+ \int_{\mathcal{Y}} \mathrm{\Psi}_2^* (-v_\phi(y)) \, \mathrm{d}\nu(y) \Bigg].
\end{multline}

To optimize this objective efficiently, we leverage the pushforward structure of the pretrained one-step generator ($p_{\mathrm{pre}} = G_{\mathrm{pre}\#} \mu_Z$). This allows us to reparameterize samples from the generated distribution through the latent variable $x_0 \sim \mu_Z$, where $x_1 = G_{\mathrm{pre}}(x_0)$. Applying this change of variables yields
\begin{multline} \label{eq:uot_unlearn2}
\inf_{v_\phi}\ \Bigg[ \int_{\mathcal{Z}} \mathrm{\Psi}_1^{*}\!
\Big( -\inf_{\Delta T_\theta}\big\{ c(G_{\mathrm{pre}}(x_0), (\Delta T_{\theta} \circ G_{\mathrm{pre}})(x_0)) \\
- v_\phi( (\Delta T_\theta \circ G_{\mathrm{pre}})(x_0) ) \big\} \Big) \mathrm{d} \mu_{Z}(x_0) 
+ \int_{\mathcal{Y}} \mathrm{\Psi}_2^* (-v_\phi(y)) \, \mathrm{d}\nu(y) \Bigg].
\end{multline}

By identifying the composed mapping $(\Delta T_\theta \circ G_{\mathrm{pre}})$ with the fine-tuned generator $G_{\theta}$, we obtain the following unlearning objective:
\begin{multline} \label{eq:unlearning_objective1}
\inf_{v_\phi}\ \Bigg[ \int_{\mathcal{Z}} \mathrm{\mathrm{\Psi}}_1^{*}\!\Big( -\inf_{G_\theta}\big\{ c_{\mathrm{ul}} (G_{\mathrm{pre}}(x_0), G_{\theta}(x_0)) - v_\phi(G_\theta(x_0)) \big\} \Big) \mathrm{d}\mu_Z(x_0) \\
+ \int_{\mathcal{Y}} \mathrm{\mathrm{\Psi}}_2^{*}\!\big( -v_\phi(y) \big) \mathrm{d}\nu(y) \Bigg].
\end{multline}

Finally, in our constrained unlearning setting where real retain data cannot be accessed, directly evaluating the expectation with respect to the data distribution $\nu=p_{\mathrm{data}}$ is infeasible. To address this issue, we approximate the target distribution using the pretrained distribution itself, i.e., $\nu \approx p_{\mathrm{pre}}$. This approximation enables a fully data-free optimization procedure while preserving the structural guidance provided by the original generator.
\begin{multline} \label{eq:unlearning_objective2}
\inf_{v_\phi}\ \Bigg[ \int_{\mathcal{Z}} \mathrm{\mathrm{\Psi}}_1^{*}\!\Big( -\inf_{G_\theta}\big\{ c_{\mathrm{ul}} (G_{\mathrm{pre}}(x_0), G_{\theta}(x_0)) - v_\phi(G_\theta(x_0)) \big\} \Big) \mathrm{d}\mu_Z(x_0) \\
+ \int_{\mathcal{Z}} \mathrm{\mathrm{\Psi}}_2^{*}\!\big( -v_\phi(G_{\mathrm{pre}}(x_0)) \big) \mathrm{d}\mu_Z(x_0) \Bigg].
\end{multline}

\subsection{Cost Design for Unlearning} \label{sec:unlearning_cost}
To implement the objective in \cref{eq:unlearning_objective2}, we introduce the unlearning cost function $c_{\mathrm{ul}}(\cdot, \cdot)$ that explicitly penalizes the generation of the forget class while preserving the remaining concepts. We first compute an anchor vector $\mu_f$ that represents the semantic center of the forget class in a feature space:
\begin{equation}
    \mu_f = \frac{1}{|\mathcal{D}_f|} \sum_{x \in \mathcal{D}_f} f(x),
\end{equation} 
where $\mathcal{D}_f$ is a set of real forget samples and $f(\cdot)$ is a pretrained feature extractor. This anchor provides a compact representation of the forget concept in the feature space.

Using this anchor, we define the forget region directly in the generated image space. 
Specifically, the \emph{active forget region} $\mathcal{R}_f \subset \mathcal{X}$ consists of generated samples whose feature representations lie within a margin $m$ of the forget anchor $\mu_{f}$ under the cosine distance:
\begin{equation}
\mathcal{R}_f =
\Big\{
x \in \mathcal{X}
\;\Big|\;
d_{\cos}\big(f(x),\mu_f\big) < m
\Big\}.
\end{equation}
where $d_{\cos}$ denotes the cosine distance and $m$ defines the semantic boundary of the forget concept in the feature space. 
During training, generated samples are dynamically checked against this region to identify forget-like outputs.

Based on this region, the unlearning cost is defined as
\begin{equation}
c_{\mathrm{ul}}(G_{\mathrm{pre}}(x_0),G_\theta(x_0)) =
\begin{cases}
\lambda \cdot \big( m - d_{\cos}(f(G_\theta(x_0)),\mu_f)\big), & \text{if } G_\theta(x_0)\in\mathcal{R}_f, \\
\tau \cdot \|G_{\mathrm{pre}}(x_0)-G_\theta(x_0)\|_2^2, & \text{otherwise}.
\end{cases}
\label{eq:unlearning_cost}
\end{equation}
This cost function provides two roles in our unlearning framework:
\begin{itemize}
    \item \textbf{Forget Cost (Active Expulsion):} For generated samples inside $\mathcal{R}_f$, a hinge-like penalty actively pushes the generated features away from the forget anchor $\mu_f$ beyond the margin $m$.
    \item \textbf{Retain Cost (Fidelity \& Transport):}
    For samples outside $\mathcal{R}_f$, we impose a squared $L_2$ cost between the current output $G_\theta(x_0)$ and the pretrained output $G_{\mathrm{pre}}(x_0)$. This term preserves the fidelity of the remaining classes while simultaneously serving as the transport cost in the UOT objective.
\end{itemize}
$\lambda$ and $\tau$ act as balancing weights for the active forgetting and the fidelity retention, respectively. Through this design, our UOT framework safely redistributes the forgotten concepts into high-quality retain samples.

\begin{algorithm}[t]
\caption{Class Unlearning via UOT}
\label{alg:uot_unlearning}
\begin{algorithmic}[1]
\Require pretrained generator $G_{\mathrm{pre}}$, unlearned generator $G_\theta$ (initialized from $G_{\mathrm{pre}}$), dual potential $v_\phi$, pre-computed forget anchor $\mu_f$.
\For{each training iteration}
  \State Sample independent noise batches $\mathcal{B}_1, \mathcal{B}_2, \mathcal{B}_3 \sim \mu_Z$
  \State Compute $c_{\mathrm{ul}} (G_{\mathrm{pre}}(x_0), G_{\theta}(x_0))$ for $x_0 \in \mathcal{B}_1 \cup \mathcal{B}_3$ via \cref{eq:unlearning_cost}
  
  \State \textbf{Update Dual Potential}
  \State Compute $\mathcal{L}_{v} = \frac{1}{|\mathcal{B}_1|}\sum_{x_0 \in \mathcal{B}_1} \mathrm{\Psi}_1^{*}\!\Big(v_\phi(G_\theta(x_0)) - c_{\mathrm{ul}} (G_{\mathrm{pre}}(x_0), G_{\theta}(x_0))\Big)$
  \State \hspace{2.2cm} $+ \frac{1}{|\mathcal{B}_2|}\sum_{x_0 \in \mathcal{B}_2} \mathrm{\Psi}_2^{*}\!\Big(-v_\phi(G_{\mathrm{pre}}(x_0))\Big)$
  \State Update $\phi$ to minimize $\mathcal{L}_{v}$
  
  \State \textbf{Update Generator}
  \State Compute $\mathcal{L}_G = \frac{1}{|\mathcal{B}_3|}\sum_{x_0 \in \mathcal{B}_3} \Big[ c_{\mathrm{ul}} (G_{\mathrm{pre}}(x_0), G_{\theta}(x_0)) - v_\phi(G_\theta(x_0)) \Big]$
  \State Update $\theta$ to minimize $\mathcal{L}_G$
\EndFor
\end{algorithmic}
\end{algorithm}

\section{Related Works}
Existing unlearning methodologies for generative models are often tailored to the iterative denoising structure of diffusion models.
As a foundational baseline, Gradient Ascent (GA) \cite{thudi2022unrolling} attempts to reverse the training process by directly applying gradient ascent on the forgetting data. However, this naive parameter-space update often leads to severe instability and catastrophic forgetting, degrading the overall generation quality. To mitigate such issues, Selective Amnesia (SA) \cite{heng2023selective} uses Elastic Weight Consolidation (EWC) to penalize parameter changes based on the Fisher Information Matrix (FIM), but its reliance on expensive FIM calculations and generative replay limits real-time efficiency. Saliency Unlearning (SalUn) \cite{fan2024salun} identifies sensitive weights through gradient-based saliency to create forgetting masks, though its performance is highly sensitive to gradient and architectural biases. Variational Diffusion Unlearning (VDU) \cite{panda2024variational} introduces a variational inference framework that balances plasticity and stability to forget specific classes, reducing computational overhead but still tied to the noise scheduling and transitional kernels of diffusion models.

Unlike these diffusion-centric approaches, our UOT-based framework (\cref{sec:method}) serves as a universal unlearning solution tailored for one-step generators. As summarized in Table~\ref{tab:data_requirements}, our structure-agnostic method overcomes the intensive data dependencies of prior works by requiring strictly zero real data during optimization, enabling a significantly more efficient unlearning procedure for one-step generators.



\begin{table}[t]
\centering
\caption{Comparison of data requirements during the \textbf{unlearning optimization phase}. Our UOT-based framework is the only method that operates with \textbf{zero real data}, relying solely on synthetic samples and a pre-computed centroid.}
\label{tab:data_requirements}
\footnotesize
\begin{tabular}{l c c c}
\toprule
\multirow{2}{*}{\textbf{Method}} & \multicolumn{3}{c}{\textbf{Optimization-phase Data}} \\
\cmidrule(lr){2-4}
 & \textbf{Real Forget} & \textbf{Real Retain} & \textbf{Generated data} \\
\midrule
Gradient Ascent (GA) & Required & None     & None     \\
VDU                  & Required & None     & None     \\
Selective Amnesia    & None     & None     & Required \\
SalUn                & Required & Required & Required$^*$ \\
\midrule
\textbf{UOT-Unlearn (Ours)} & \textbf{None}$^\dagger$ & \textbf{None} & \textbf{Required} \\
\bottomrule
\end{tabular}

{\raggedright
\scriptsize 
\noindent $^*$ SalUn utilizes generated data only in specific large-scale generation tasks. \\
\noindent $^\dagger$ Real forget data is used only for one-time pre-computation of $\mu_f$, ensuring no real data is accessed during the optimization phase. \par}
\vspace{-8pt}
\end{table}

\section{Experiments}
\label{sec:experiments}

This section provides a systematic evaluation of our UOT-based unlearning framework. Our experiments are organized to progressively validate the core properties of the proposed method. In \cref{sec:exp_2d_synthetic}, we utilize a 2D synthetic dataset to visually analyze the probability redistribution during the unlearning process. In \cref{sec:exp_image}, we evaluate the framework on high-dimensional image benchmarks, including CIFAR-10 \cite{krizhevsky2009learning} and ImageNet-256 \cite{russakovsky2015imagenet}, to assess its scalability and the fundamental trade-off between concept erasure and generative fidelity. In \cref{sec:exp_ablation}, we conduct an ablation study to characterize the sensitivity of the optimization dynamics to core hyperparameters. Detailed implementation settings, including specific network architectures and hyperparameter configurations, are provided in Appendix.

We compare our approach against the established unlearning baselines: \textit{Gradient Ascent (GA)}, \textit{Selective Amnesia (SA) \cite{heng2023selective}}, \textit{Saliency Unlearning (SalUn) \cite{fan2024salun}}, and \textit{Variational Diffusion Unlearning (VDU) \cite{panda2024variational}}. Because these methods are designed for multi-step iterative denoising, we adapt their objectives to operate within a single forward-pass generation framework (see Appendix for adaptation details).

\subsection{2D Synthetic Data} \label{sec:exp_2d_synthetic}
We utilize a 2D synthetic dataset consisting of three Gaussian modes to examine the unlearning dynamics in a visually interpretable setting. The primary goal of this experiment is to evaluate whether the probability associated with the forget target is successfully re-assigned to the remaining classes.

\begin{figure}[tb]
  \centering
  \begin{subfigure}{0.32\linewidth}
    \centering
    \includegraphics[width=\linewidth]{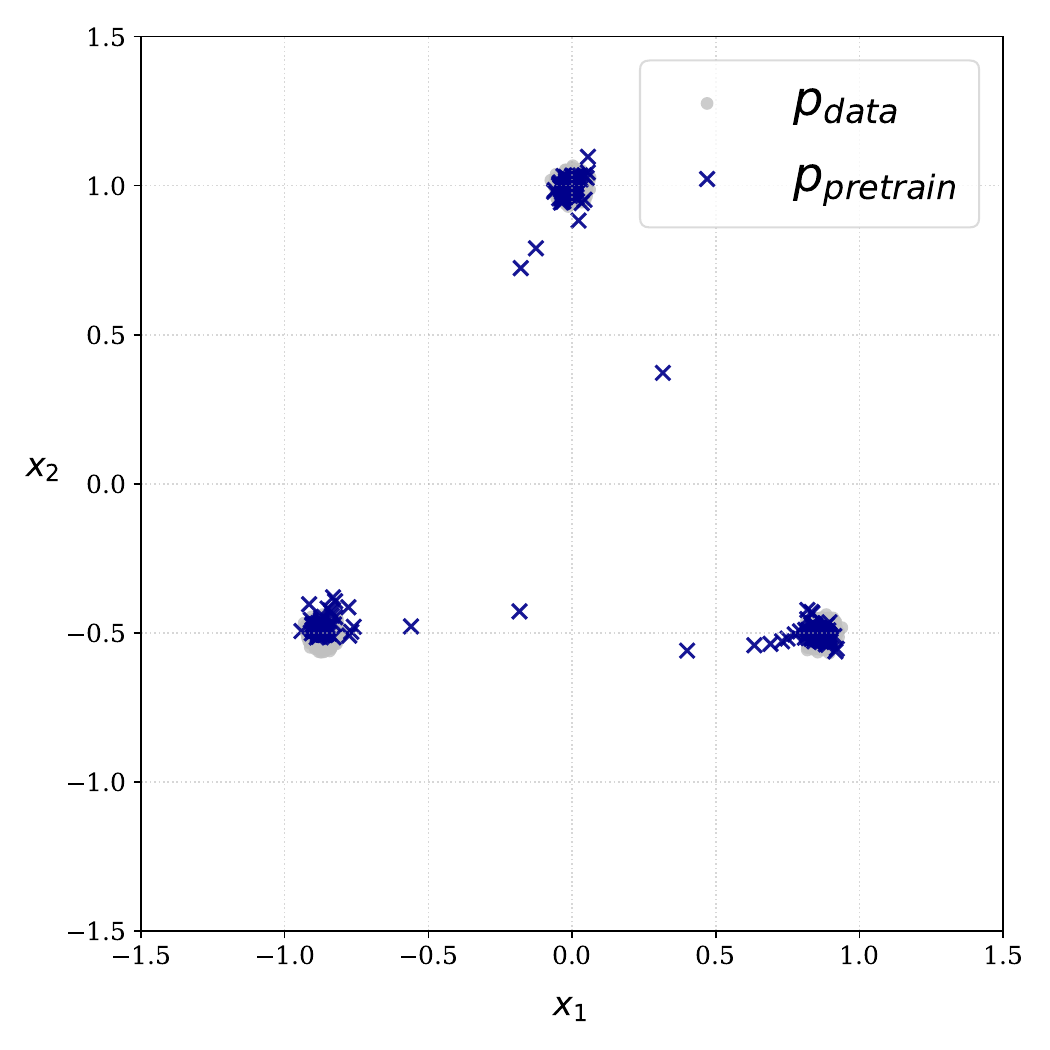}
    \caption{Pretrained}
    \label{fig:toy_pretrained}
  \end{subfigure}
  \hfill
  \begin{subfigure}{0.32\linewidth}
    \centering
    \includegraphics[width=\linewidth]{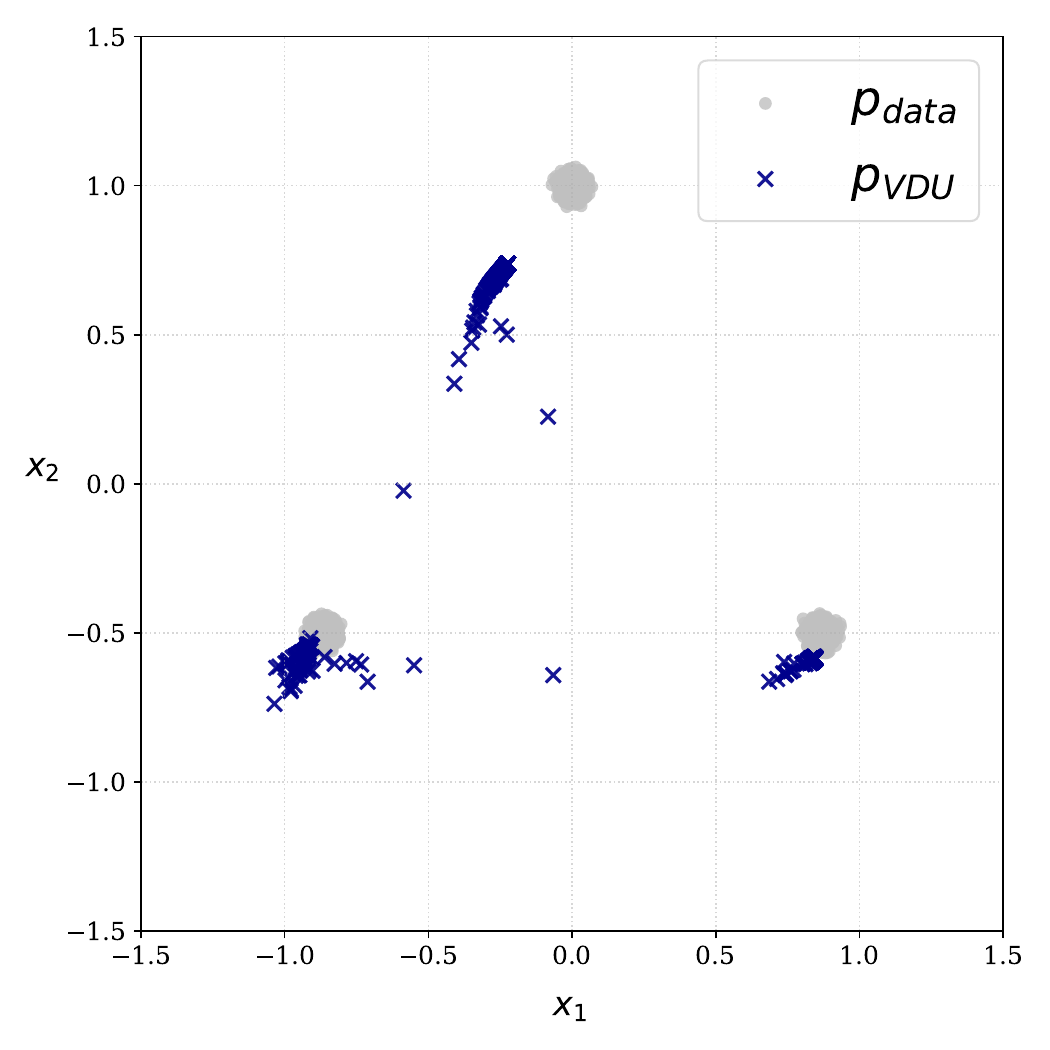}
    \caption{VDU (Baseline)}
    \label{fig:toy_vdu}
  \end{subfigure}
  \hfill
  \begin{subfigure}{0.32\linewidth}
    \centering
    \includegraphics[width=\linewidth]{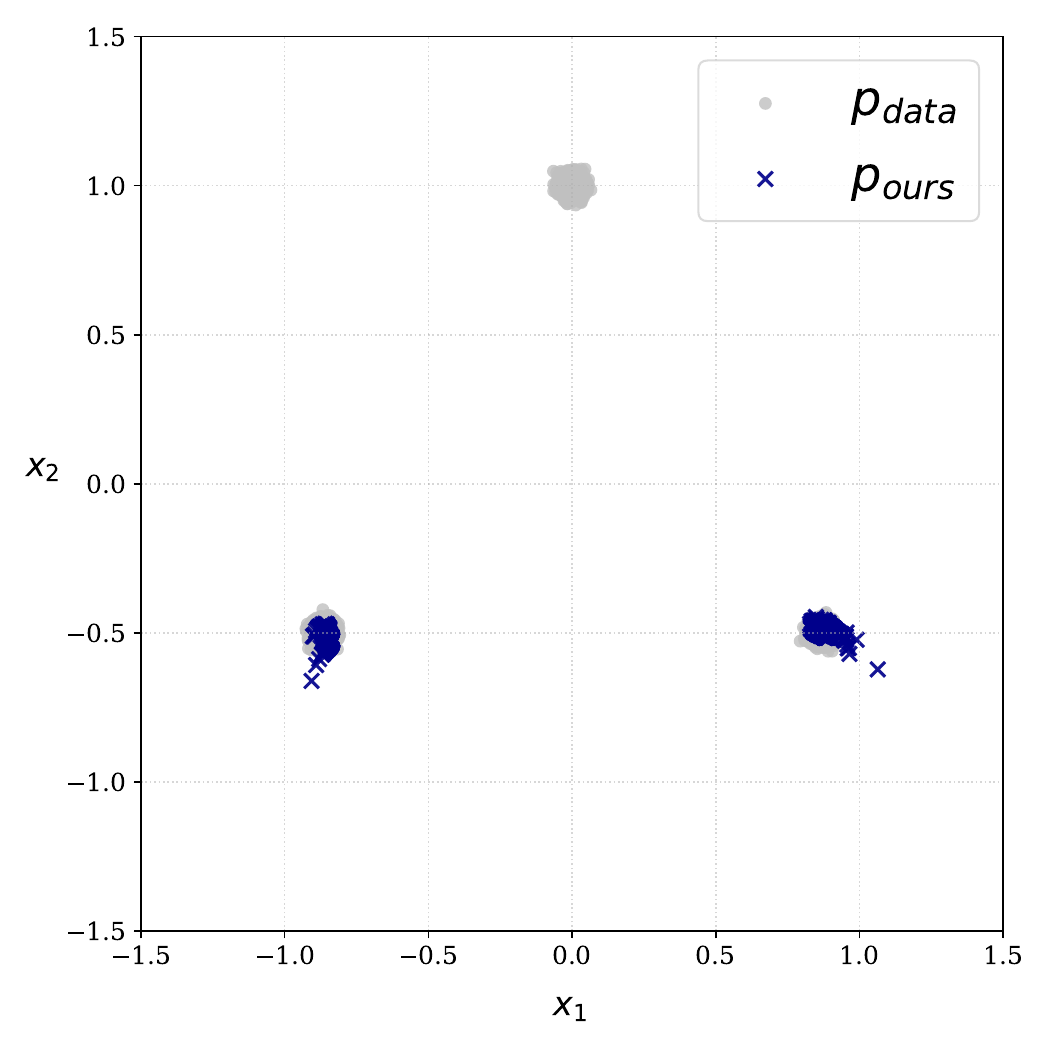}
    \caption{UOT-Unlearn (Ours)}
    \label{fig:toy_ours}
  \end{subfigure}  
  \caption{\textbf{Unlearning results on a 2D toy dataset} where the forget mode is located at $(0,1)$. (a) Pretrained one-step generator. (b) VDU leads to overall distribution distortion. (c) Our method redistributes the forget mode to the remaining modes.}
  \label{fig:toy_experiment}
  \vspace{-8pt}
\end{figure}

Our UOT-based framework enables a principled redistribution of probability mass by leveraging the $f$-divergence penalty to relax strict marginal constraints. As shown in \cref{fig:toy_experiment}, the probability density previously assigned to the forget mode is smoothly remapped toward the supports of the retain modes.
This mechanism ensures that the generator remains within the support of the remaining classes, effectively displacing the targeted concept while preserving the shape and density of the retain distribution. In contrast, while the baseline method  (VDU \cite{panda2024variational}) is successful in unlearning the forget mode (at $(0, 1)$), the removed probability mass is redistributed into the invalid regions outside the support of $p_{data}$.

\subsection{Image Unlearning Benchmarks} \label{sec:exp_image}
\paragraph{Experimental Setup.}
We evaluate our method on image-generation benchmarks. Our primary experiments are conducted on the CIFAR-10 dataset using Consistency Trajectory Models (CTM) \cite{kim2023consistency} and MeanFlow (MF) \cite{geng2025mean} as representative one-step generative architectures.
To assess concept erasure performance, we perform single-class unlearning targeting classes 1 (\textit{automobile}), 6 (\textit{frog}), and 8 (\textit{ship}). 
We further scale UOT-Unlearn to ImageNet-256 using the class-conditional Meanflow model. To apply our unconditional unlearning framework, we marginalize over class labels, effectively treating it as an unconditional generator. We focus on aquatic classes to evaluate semantic shift.
To compute the unlearning cost $c_{\mathrm{ul}}$, we employ pretrained networks as the feature extractor $f(\cdot)$. In constructing the forget anchor $\mu_f$, we use only a small subset of forget-class samples (512 images), demonstrating that the proposed method requires minimal data to identify the target concept.

\begin{figure}[tb]
  \centering
  \small 
  \begin{tabular}{@{} c @{\hskip 8pt} ccc @{\hskip 5pt}c@{\hskip 5pt} ccc @{}}
    & \multicolumn{3}{c}{\kern-2pt Pretrained\kern2pt} & & \multicolumn{3}{c}{Unlearned} \\[-2pt]
    
    \raisebox{1.5ex}{\rotatebox[origin=c]{90}{CTM}} &
    \begin{subfigure}[c]{0.14\linewidth}
      \includegraphics[width=\linewidth]{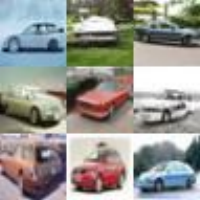}
      \caption{Class 1}
    \end{subfigure} &
    \begin{subfigure}[c]{0.14\linewidth}
      \includegraphics[width=\linewidth]{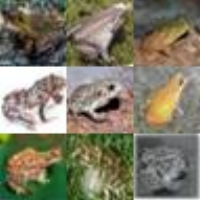}
      \caption{Class 6}
    \end{subfigure} &
    \begin{subfigure}[c]{0.14\linewidth}
      \includegraphics[width=\linewidth]{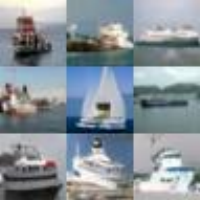}
      \caption{Class 8}
    \end{subfigure} &
    \raisebox{1.5ex}{%
      \begin{tikzpicture}[baseline=(current bounding box.center)]
        \draw[dash pattern=on 3pt off 2pt, gray!80, thick] (0,-1.1) -- (0,1.1);
      \end{tikzpicture}%
    } &
    \begin{subfigure}[c]{0.14\linewidth}
      \includegraphics[width=\linewidth]{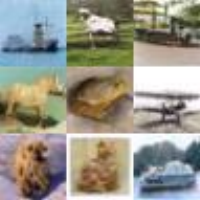}
      \caption{Class 1}
    \end{subfigure} &
    \begin{subfigure}[c]{0.14\linewidth}
      \includegraphics[width=\linewidth]{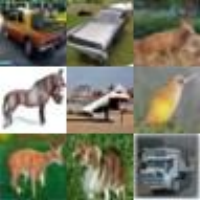}
      \caption{Class 6}
    \end{subfigure} &
    \begin{subfigure}[c]{0.14\linewidth}
      \includegraphics[width=\linewidth]{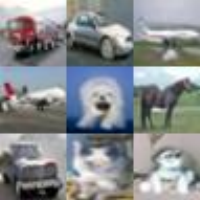}
      \caption{Class 8}
    \end{subfigure} \\

    \cmidrule(l{2pt}r{6pt}){2-4} \cmidrule(l{3pt}r{2pt}){6-8} 

    \raisebox{1.5ex}{\rotatebox[origin=c]{90}{MF}} &
    \begin{subfigure}[c]{0.14\linewidth}
      \includegraphics[width=\linewidth]{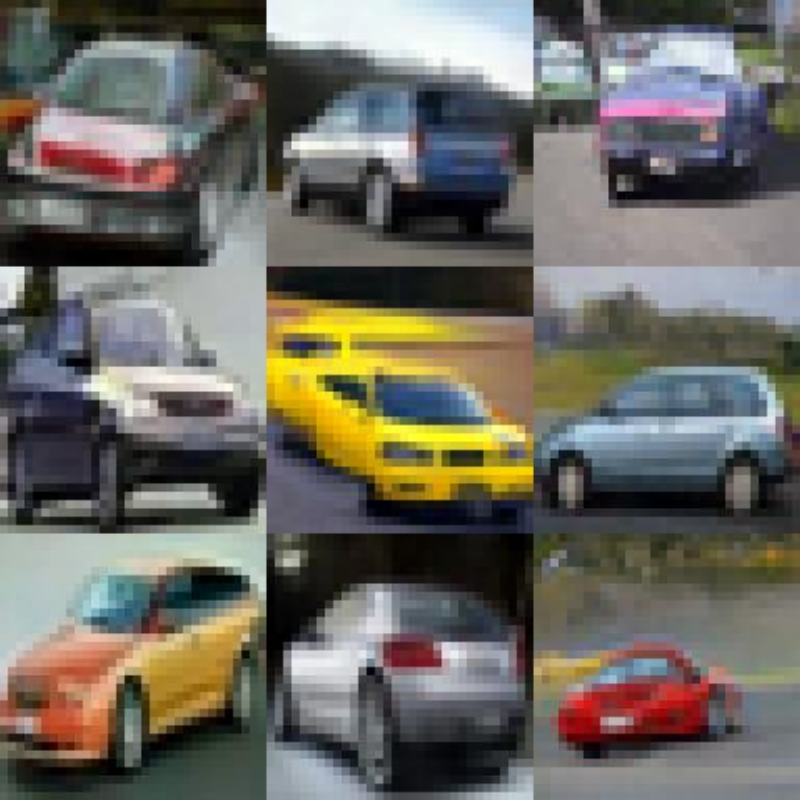}
      \caption{Class 1}
    \end{subfigure} &
    \begin{subfigure}[c]{0.14\linewidth}
      \includegraphics[width=\linewidth]{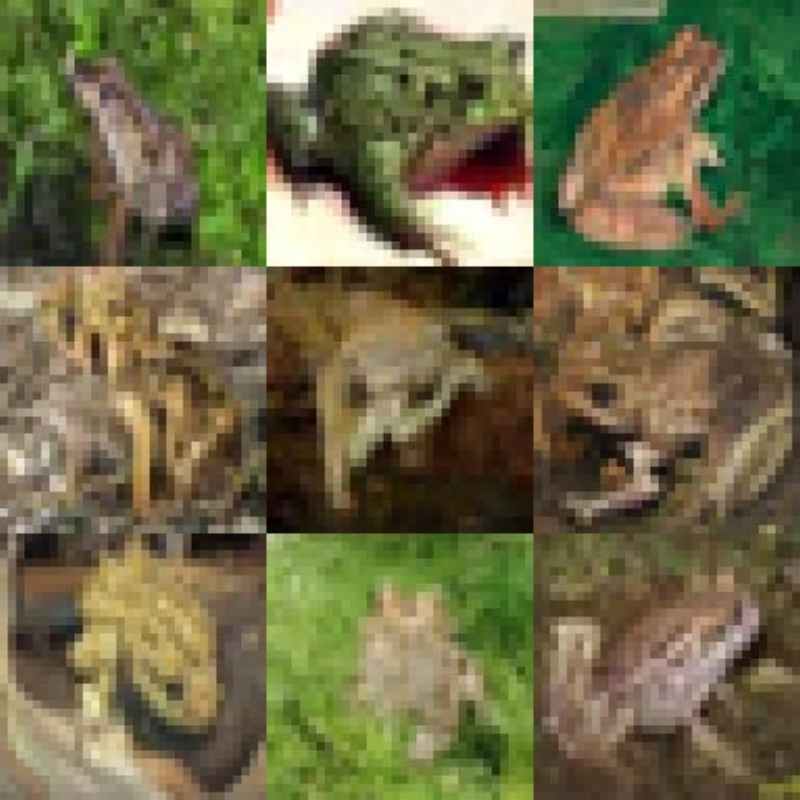}
      \caption{Class 6}
    \end{subfigure} &
    \begin{subfigure}[c]{0.14\linewidth}
      \includegraphics[width=\linewidth]{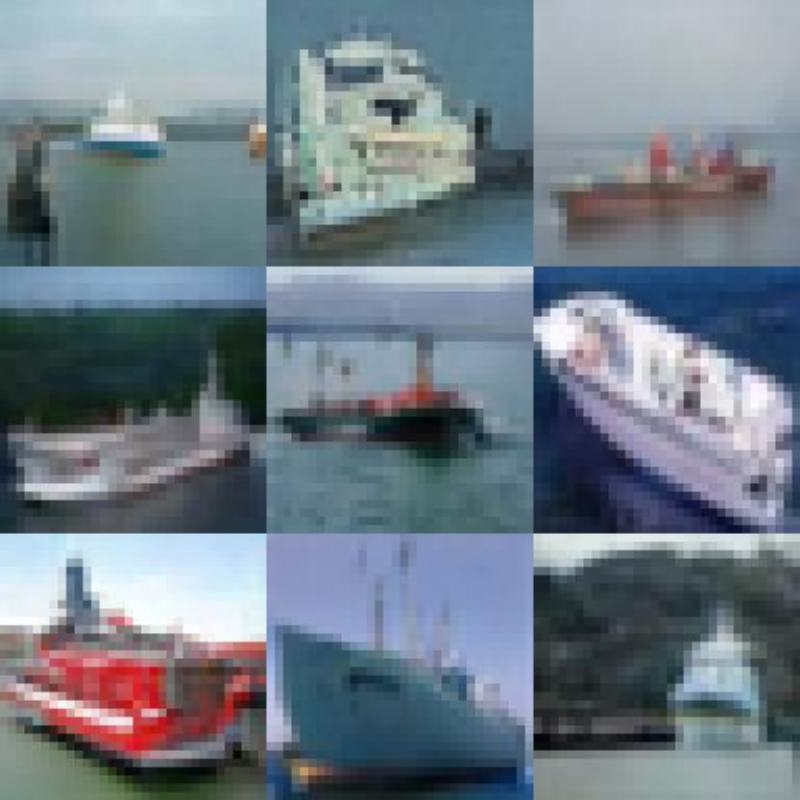}
      \caption{Class 8}
    \end{subfigure} &
    \raisebox{1.5ex}{%
      \begin{tikzpicture}[baseline=(current bounding box.center)]
        \draw[dash pattern=on 3pt off 2pt, gray!80, thick] (0,-1.1) -- (0,1.1);
      \end{tikzpicture}%
    } &
    \begin{subfigure}[c]{0.14\linewidth}
      \includegraphics[width=\linewidth]{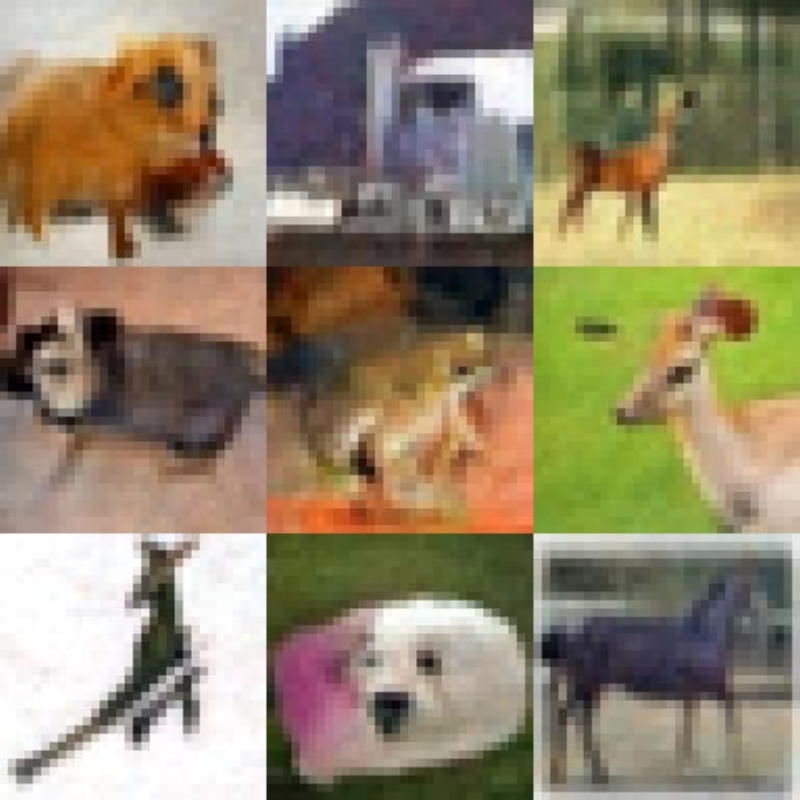}
      \caption{Class 1}
    \end{subfigure} &
    \begin{subfigure}[c]{0.14\linewidth}
      \includegraphics[width=\linewidth]{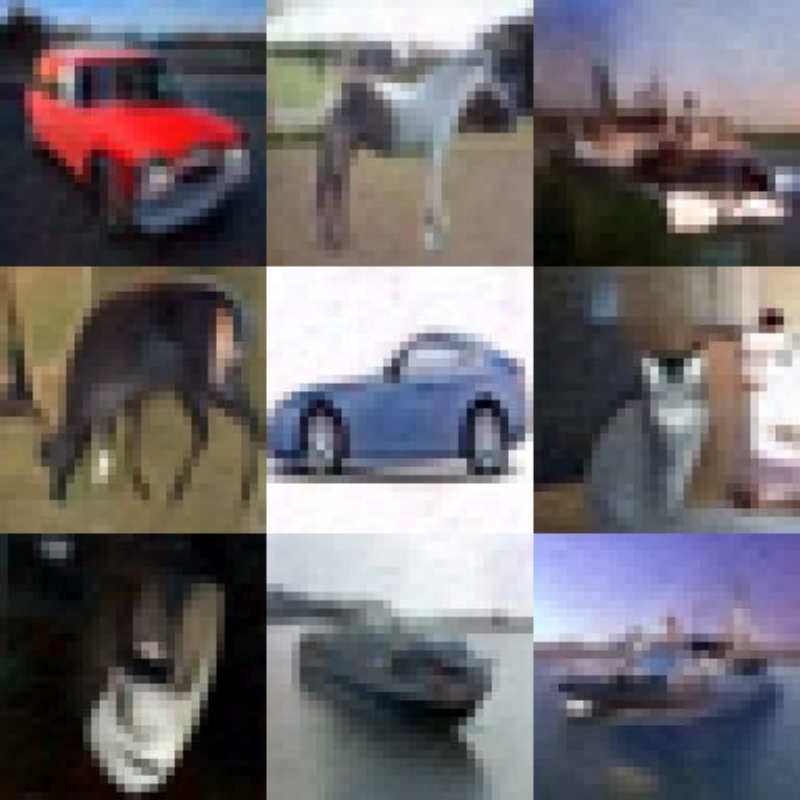}
      \caption{Class 6}
    \end{subfigure} &
    \begin{subfigure}[c]{0.14\linewidth}
      \includegraphics[width=\linewidth]{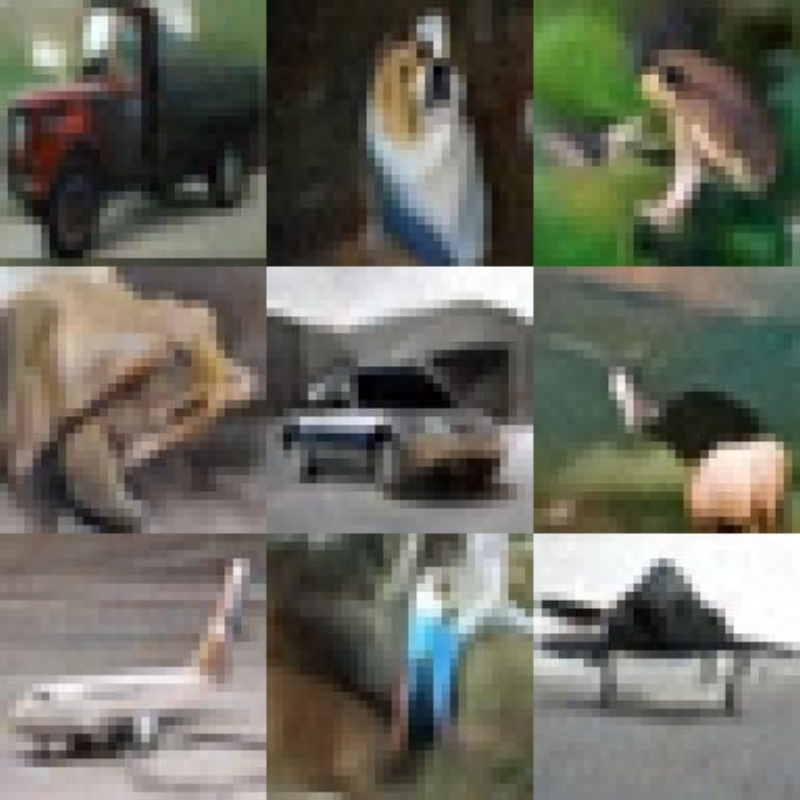}
      \caption{Class 8}
    \end{subfigure}
  \end{tabular}
  \caption{\textbf{Qualitative unlearning results on CIFAR-10}. Unconditional samples of target classes (1, 6, and 8) generated by CTM (top) and MF (bottom) architectures. To clearly illustrate the semantic erasure of targeted concepts, the \textit{Unlearned} outputs are generated using the exact same initial noise seeds as their \textit{Pretrained} counterparts.}
  \label{fig:qualitative_results}
  \vspace{-8pt}
\end{figure}
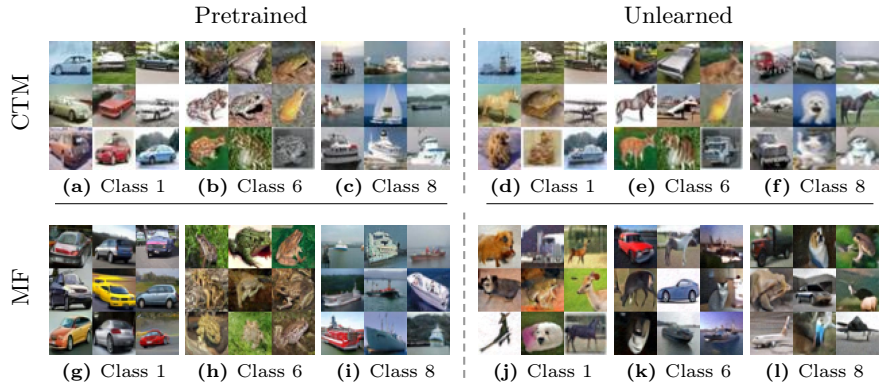

\paragraph{Evaluation Metrics.}
To quantitatively evaluate the performance, we employ two primary metrics. First, \textbf{Percentage of Unlearning (PUL) \cite{tiwary2025adapt}} measures the effectiveness of removing the forget class by computing the relative reduction in its generation frequency.
Let $N_{\mathrm{pre}}$ and $N_{\mathrm{unl}}$ denote the number of images generated by the pretrained and unlearned models, respectively, that are classified as the forget class.  The PUL score is defined as
\begin{equation}
    \text{PUL} = \frac{N_{\mathrm{pre}} - N_{\mathrm{unl}}}{N_{\mathrm{pre}}} \times 100 \%.
\end{equation}
To ensure fair evaluation, we use independent classifiers that are not involved in the training procedure, i.e., in computing the unlearning cost.

Second, \textbf{Unlearned FID (u-FID) \cite{panda2024variational}} evaluates the generative quality of the retained classes.
We compute the Fréchet Inception Distance (FID) \cite{heusel2017gans} between the generated image distributions and the real images restricted to the retain classes. Specifically, on CIFAR-10, we calculate the u-FID between the 45,000 real retain images (the full training set excluding the forget class) and 45,000 images randomly generated by the unlearned model. On ImageNet-256, u-FID is computed against a localized retain subset of aquatic classes to precisely quantify semantic collateral damage.

\begin{table}[t!]
  \caption{\textbf{Unlearning performance on CTM and Meanflow.} Best results are highlighted in \textbf{bold}, and second-best results are \underline{underlined}. UOT-Unlearn consistently achieves the highest Percentage of Unlearning (PUL) while preserving the original FID.}
  \label{tab:main_results}
  \centering
  \resizebox{\linewidth}{!}{
  \begin{tabular}{@{}l c cc cc cc cc cc@{}}
    \toprule
    \textbf{Unlearned} & \textbf{Original} & \multicolumn{2}{c}{\textbf{GA}} & \multicolumn{2}{c}{\textbf{SA}} & \multicolumn{2}{c}{\textbf{SalUn}} & \multicolumn{2}{c}{\textbf{VDU}} & \multicolumn{2}{c}{\textbf{UOT-Unlearn (Ours)}} \\
    \cmidrule(lr){3-4} \cmidrule(lr){5-6} \cmidrule(lr){7-8} \cmidrule(lr){9-10} \cmidrule(lr){11-12}
    \textbf{Class} & \textbf{FID $\downarrow$} & PUL (\%) $\uparrow$ & u-FID $\downarrow$ & PUL (\%) $\uparrow$ & u-FID $\downarrow$ & PUL (\%) $\uparrow$ & u-FID $\downarrow$ & PUL (\%) $\uparrow$ & u-FID $\downarrow$ & PUL (\%) $\uparrow$ & u-FID $\downarrow$ \\
    
    \midrule
    \multicolumn{12}{c}{Target Model: Consistency Trajectory Models (CTM) \cite{kim2023consistency}} \\
    \midrule
    Class 1 (Auto) & 4.53 & 88.07 & 160.03 & \textbf{91.16} & 49.80 & \underline{90.64} & \underline{41.36} & 53.86 & 71.98 & 80.32 & \textbf{9.90} \\
    Class 6 (Frog) & 5.02 & 39.16 & 72.17  & \underline{61.85} & 42.54 & 43.16 & \underline{39.00} & 51.25 & 47.11 & \textbf{90.98} & \textbf{5.11} \\
    Class 8 (Ship) & 4.36 & \textbf{95.40} & 208.80 & 64.50 & 62.48 & 48.65 & \underline{45.70} & 48.43 & 52.07 & \underline{85.23} & \textbf{5.88} \\
    \midrule
    \textbf{Average} & 4.64 & \underline{74.21} & 147.00 & 72.50 & 51.61 & 60.82 & \underline{42.02} & 51.18 & 57.05 & \textbf{85.51} & \textbf{6.96} \\
    
    \midrule
    \multicolumn{12}{c}{Target Model: Meanflow \cite{heng2023selective}} \\
    \midrule
    Class 1 (Auto) & 7.73 & 93.43 & 115.59 & \underline{98.05} & 82.92 & \textbf{99.35} & \underline{62.77} & 33.70 & 73.73 & 90.63 & \textbf{17.69} \\
    Class 6 (Frog) & 9.23 & \underline{60.59} & 47.40  & 58.49 & 21.14 & 50.14 & \underline{20.06} & 50.42 & 66.86 & \textbf{96.25} & \textbf{19.58} \\
    Class 8 (Ship) & 7.08 & \underline{85.74} & 49.08  & 84.98 & \underline{20.48} & 78.01 & \textbf{15.32} & 71.47 & 33.06 & \textbf{90.43} & 21.31 \\
    \midrule
    \textbf{Average} & 8.01 & 79.92 & 70.69  & \underline{80.51} & 41.51 & 75.83 & \underline{32.72} & 51.86 & 57.88 & \textbf{92.44} & \textbf{19.53} \\
    \bottomrule
  \end{tabular}
  }
\end{table}

\begin{figure}[t!]
  \centering
  \small 
  \begin{tabular}{@{} c @{\hskip 12pt} c @{\hskip 0pt} c c c @{\hskip 4pt} c @{}}
    & & \kern 13pt Class 1 & \kern 13pt Class 6 & \kern 13pt Class 8 & \\
    \noalign{\smallskip}
    $\vcenter{\hbox{\rotatebox[origin=c]{90}{CTM}}}$ & 
    $\vcenter{\hbox{\rotatebox[origin=c]{90}{\tiny u-FID $\downarrow$}}}$ &
    $\vcenter{\hbox{\includegraphics[width=0.25\linewidth]{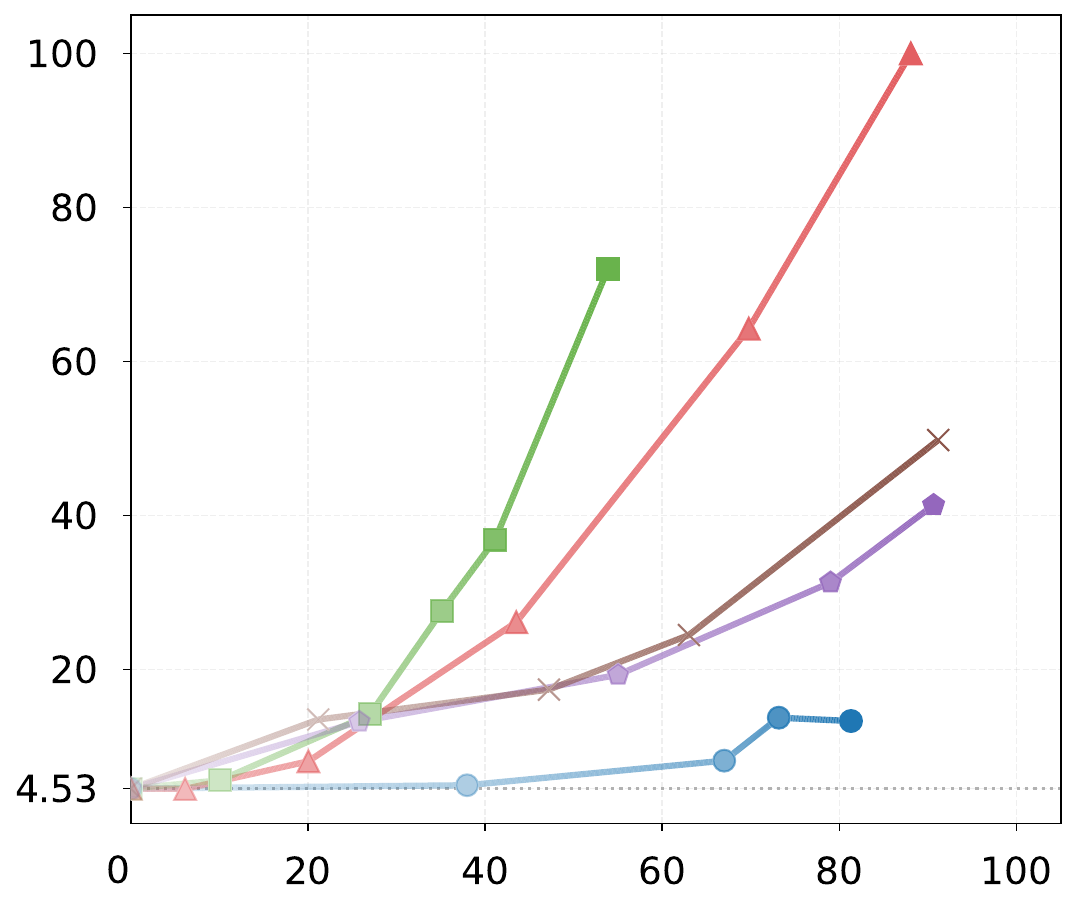}}}$ &
    $\vcenter{\hbox{\includegraphics[width=0.25\linewidth]{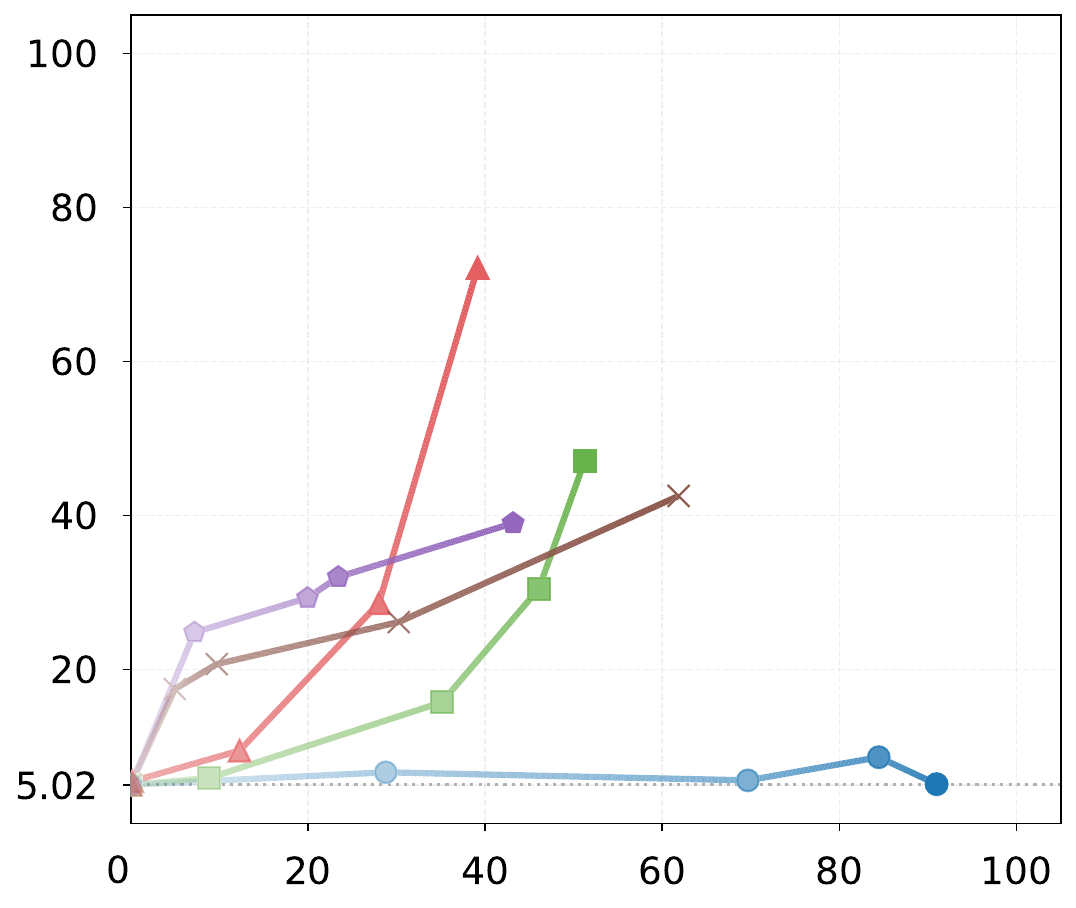}}}$ &
    $\vcenter{\hbox{\includegraphics[width=0.25\linewidth]{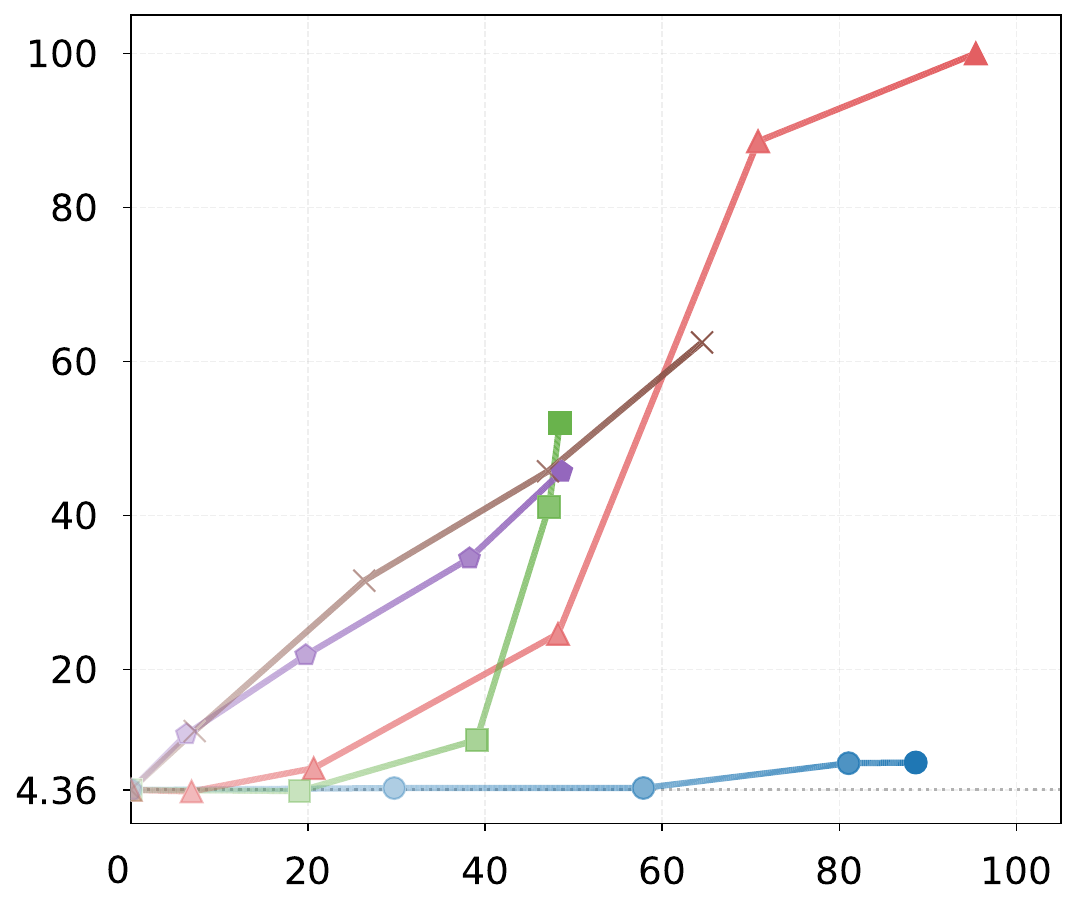}}}$ &
    $\vcenter{\hbox{\includegraphics[width=0.14\linewidth]{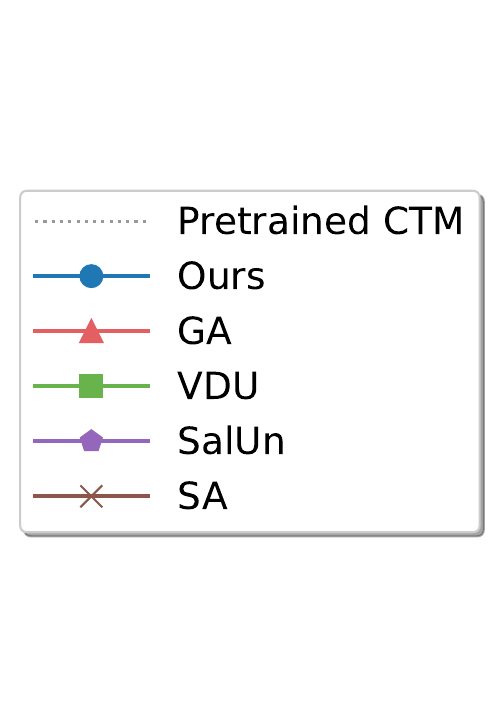}}}$ \\[0.5ex]
    & & \kern 15pt \tiny PUL (\%) $\uparrow$ & \kern 15pt \tiny PUL (\%) $\uparrow$ & \kern 15pt \tiny PUL (\%) $\uparrow$ & \\
    \noalign{\bigskip}
    $\vcenter{\hbox{\rotatebox[origin=c]{90}{MF}}}$ & 
    $\vcenter{\hbox{\rotatebox[origin=c]{90}{\tiny u-FID $\downarrow$}}}$ &
    $\vcenter{\hbox{\includegraphics[width=0.25\linewidth]{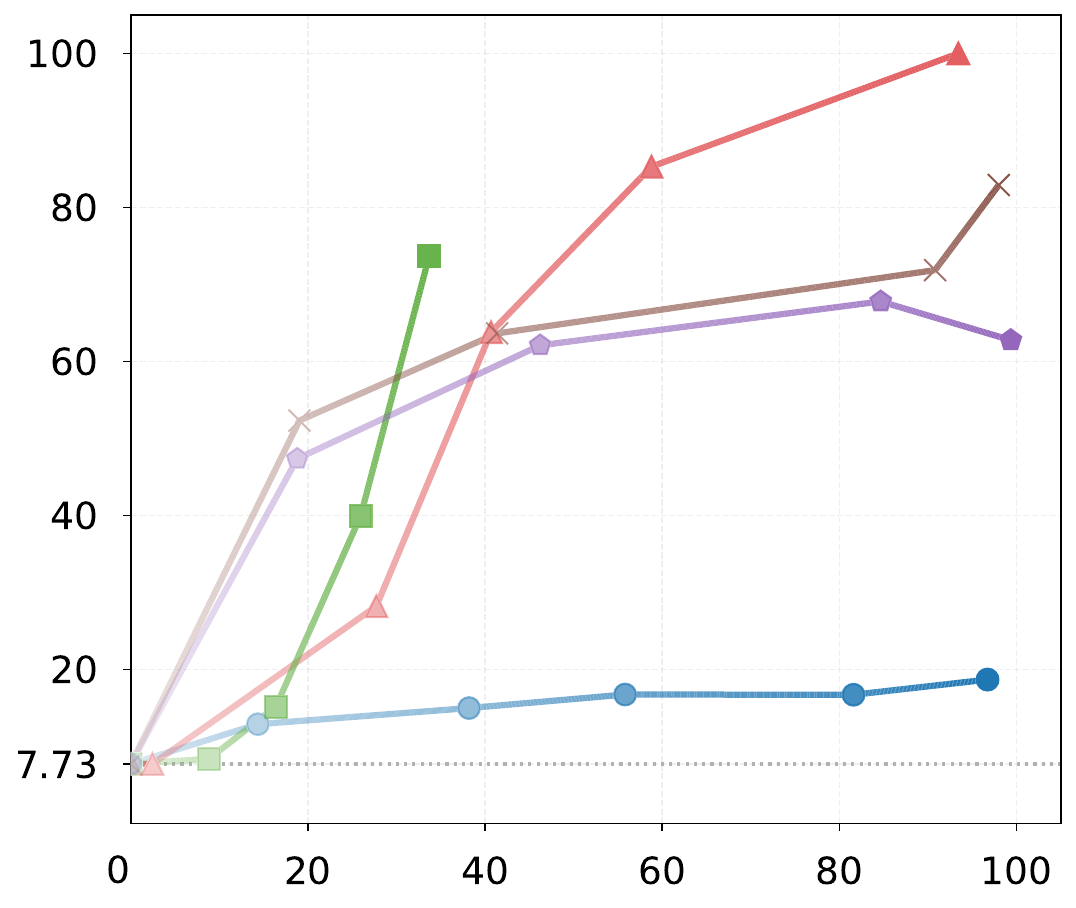}}}$ &
    $\vcenter{\hbox{\includegraphics[width=0.25\linewidth]{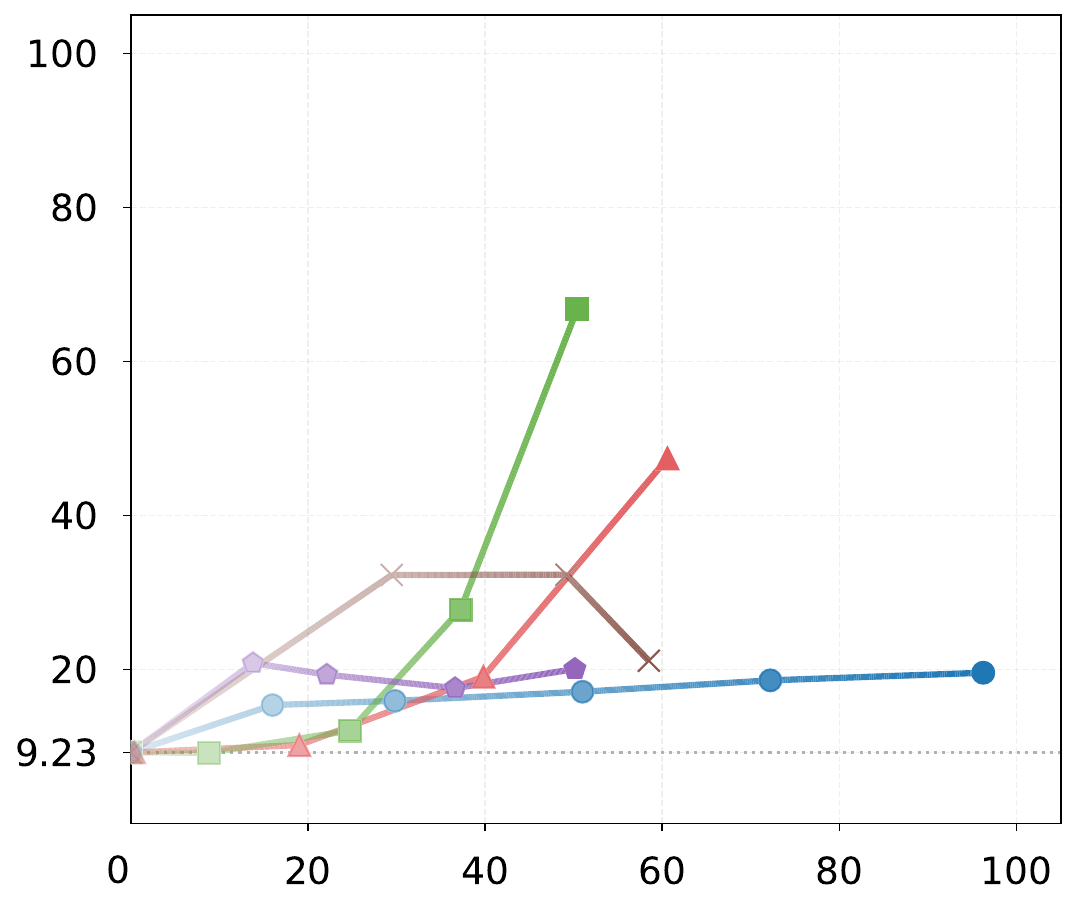}}}$ &
    $\vcenter{\hbox{\includegraphics[width=0.25\linewidth]{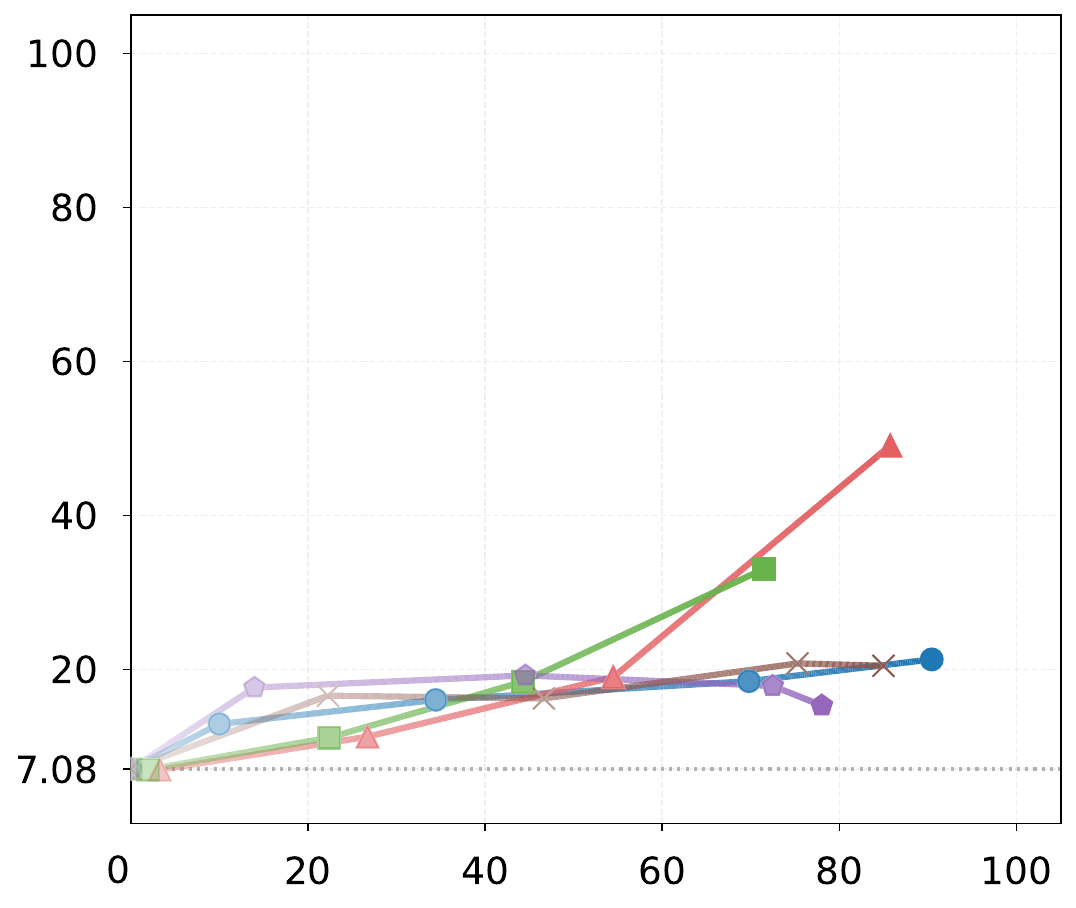}}}$ &
    $\vcenter{\hbox{\includegraphics[width=0.14\linewidth]{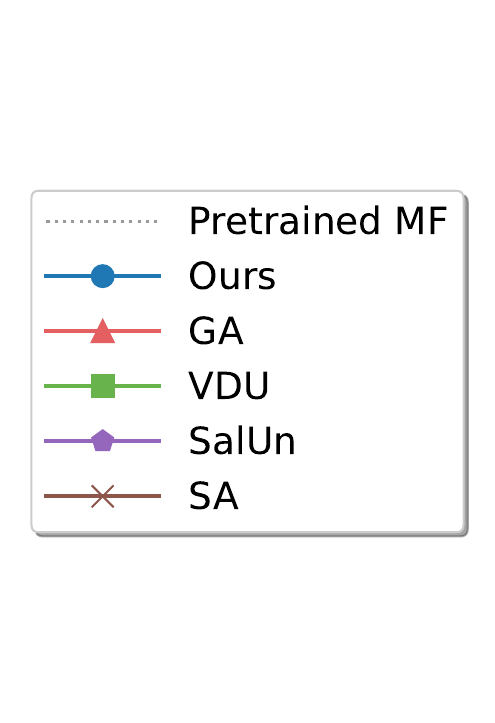}}}$ \\[0.5ex]
    & & \kern 15pt \tiny PUL (\%) $\uparrow$ & \kern 15pt \tiny PUL (\%) $\uparrow$ & \kern 15pt \tiny PUL (\%) $\uparrow$ & \\
  \end{tabular}

  \caption{\textbf{Step-wise unlearning trajectories on CIFAR-10.} We analyze the trade-off between concept erasure (PUL $\uparrow$) and generative fidelity (u-FID $\downarrow$) across unconditional CTM (top row) and MF (bottom row) models. Our framework achieves a superior trade-off, erasing target class while preserving distributional consistency. Metrics are computed sequentially using fixed initial noise seeds for fair variance analysis.}
  \label{fig:quantitative_results}
  \vspace{-8pt}
\end{figure}

\begin{figure}[t] 
  \centering
  \small 
  \setlength{\tabcolsep}{3pt} 
  
  \begin{tabular}{@{} c ccc @{}}
    & \textbf{Pretrained} & \textbf{Baseline (GA)} & \textbf{Ours}  \\
    \addlinespace[3pt] 
    \rotatebox[origin=c]{90}{\textbf{Forget Class}} &
    \adjustbox{valign=m}{\includegraphics[width=0.28\linewidth]{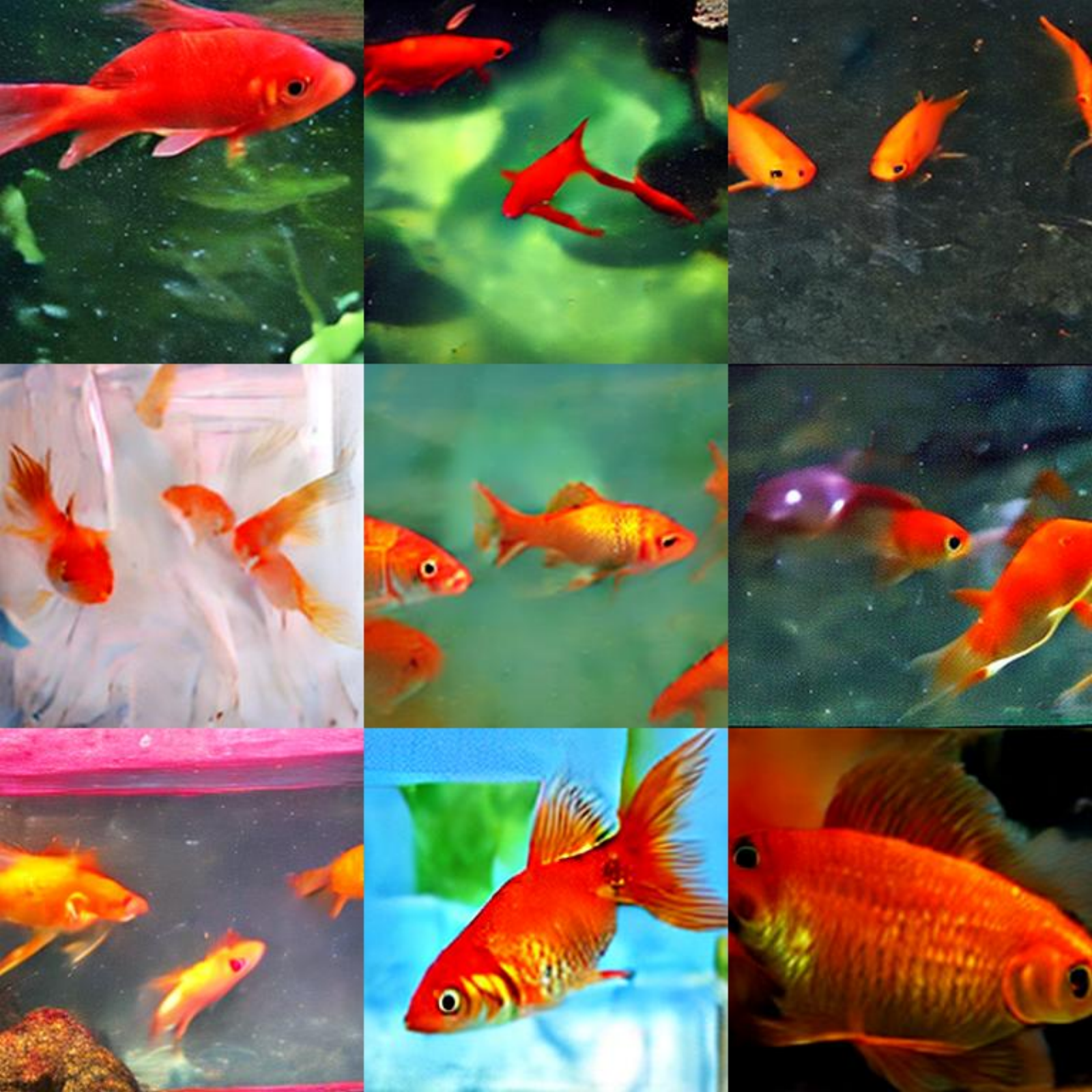}} &
    \adjustbox{valign=m}{\includegraphics[width=0.28\linewidth]{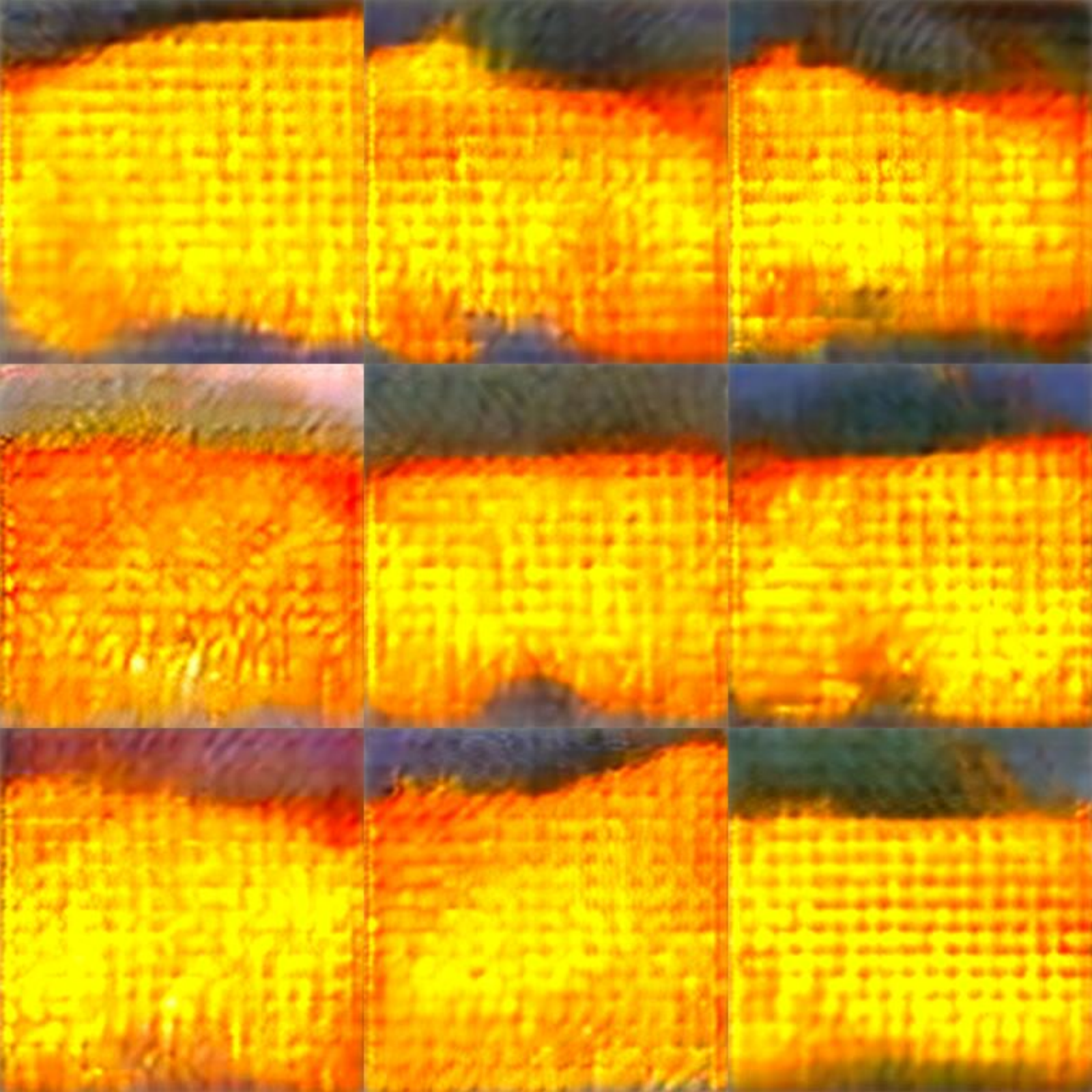}}  &
    \adjustbox{valign=m}{\includegraphics[width=0.28\linewidth]{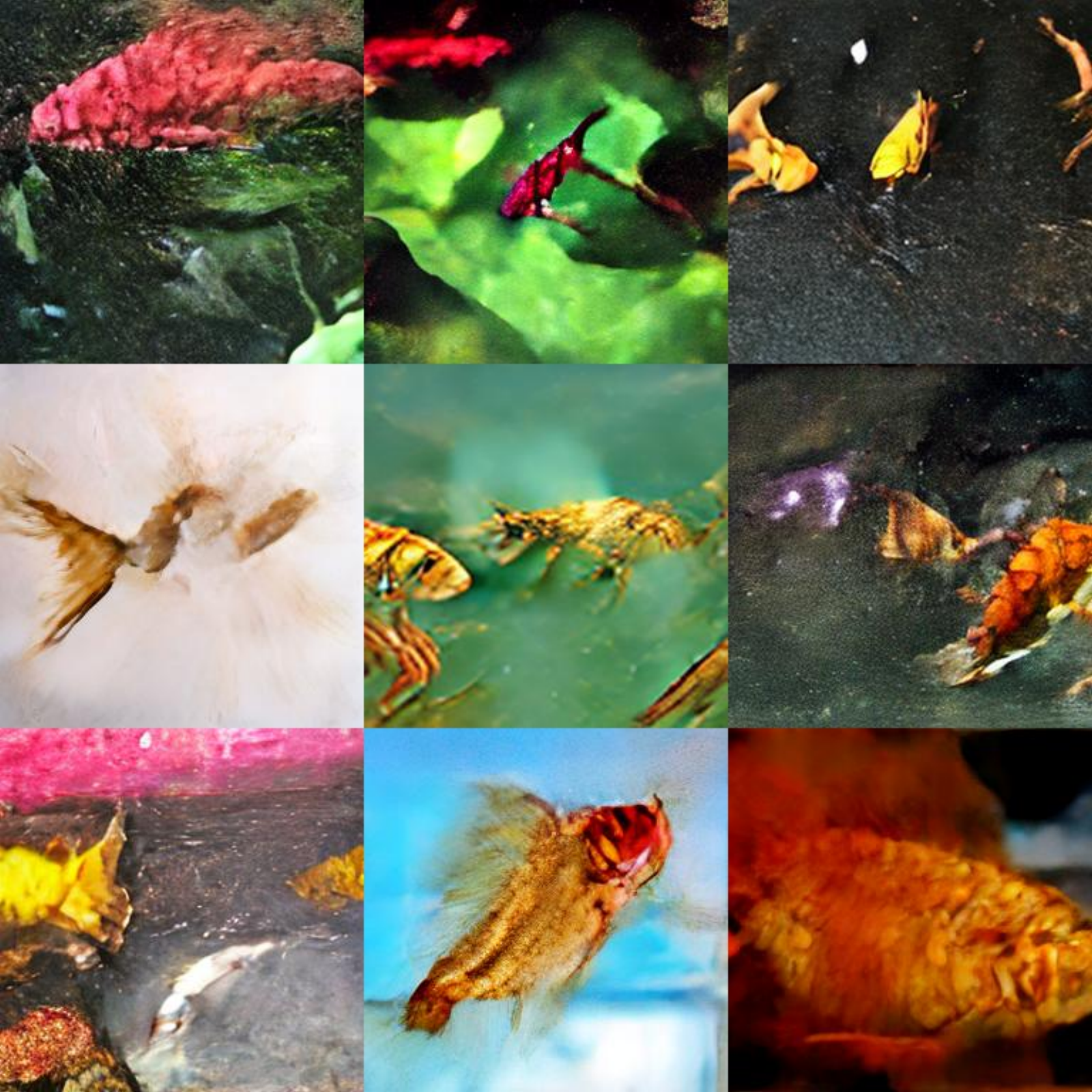}} \\
    \addlinespace[4pt] 

    \rotatebox[origin=c]{90}{\textbf{Retain Class}} &
    \adjustbox{valign=m}{\includegraphics[width=0.28\linewidth]{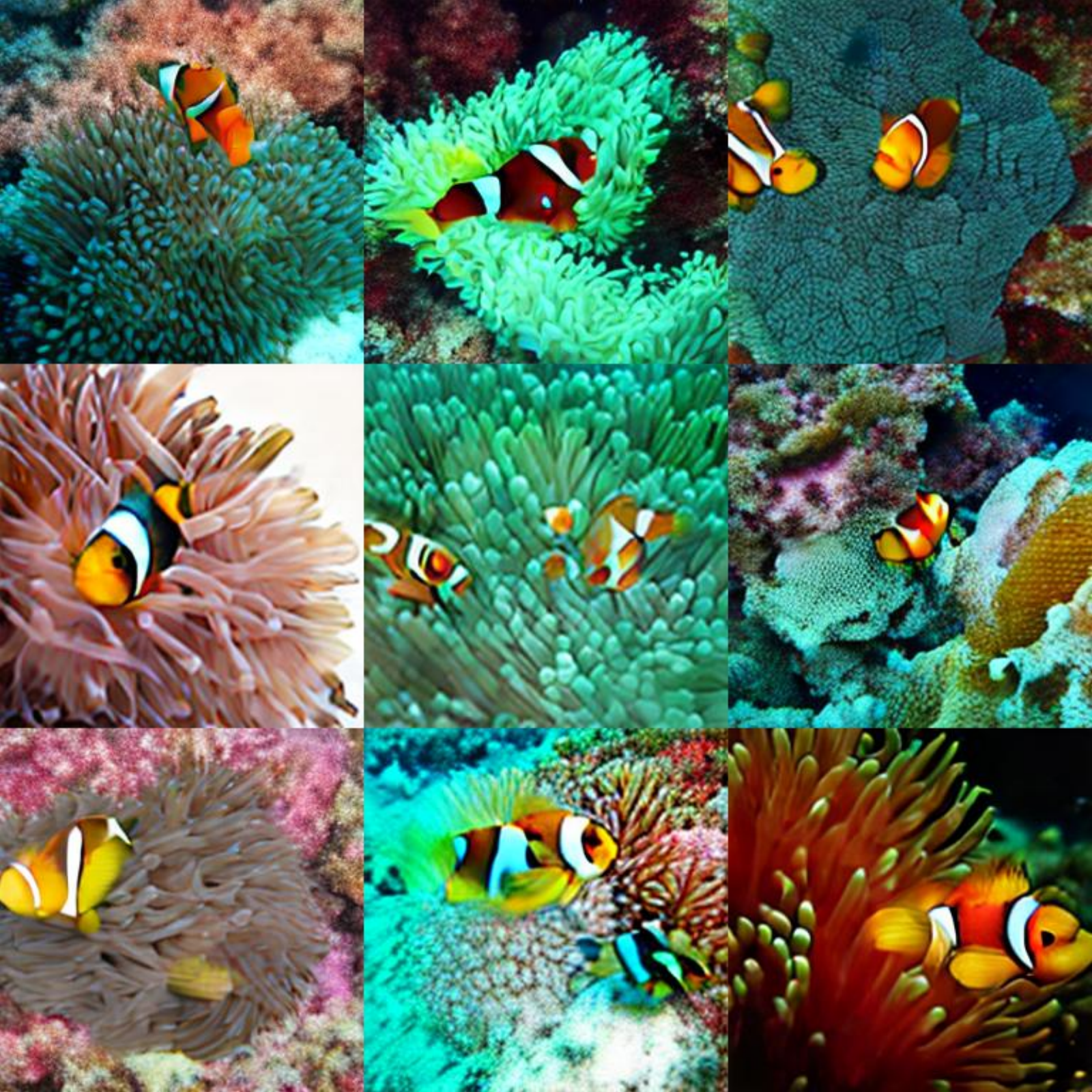}} &
    \adjustbox{valign=m}{\includegraphics[width=0.28\linewidth]{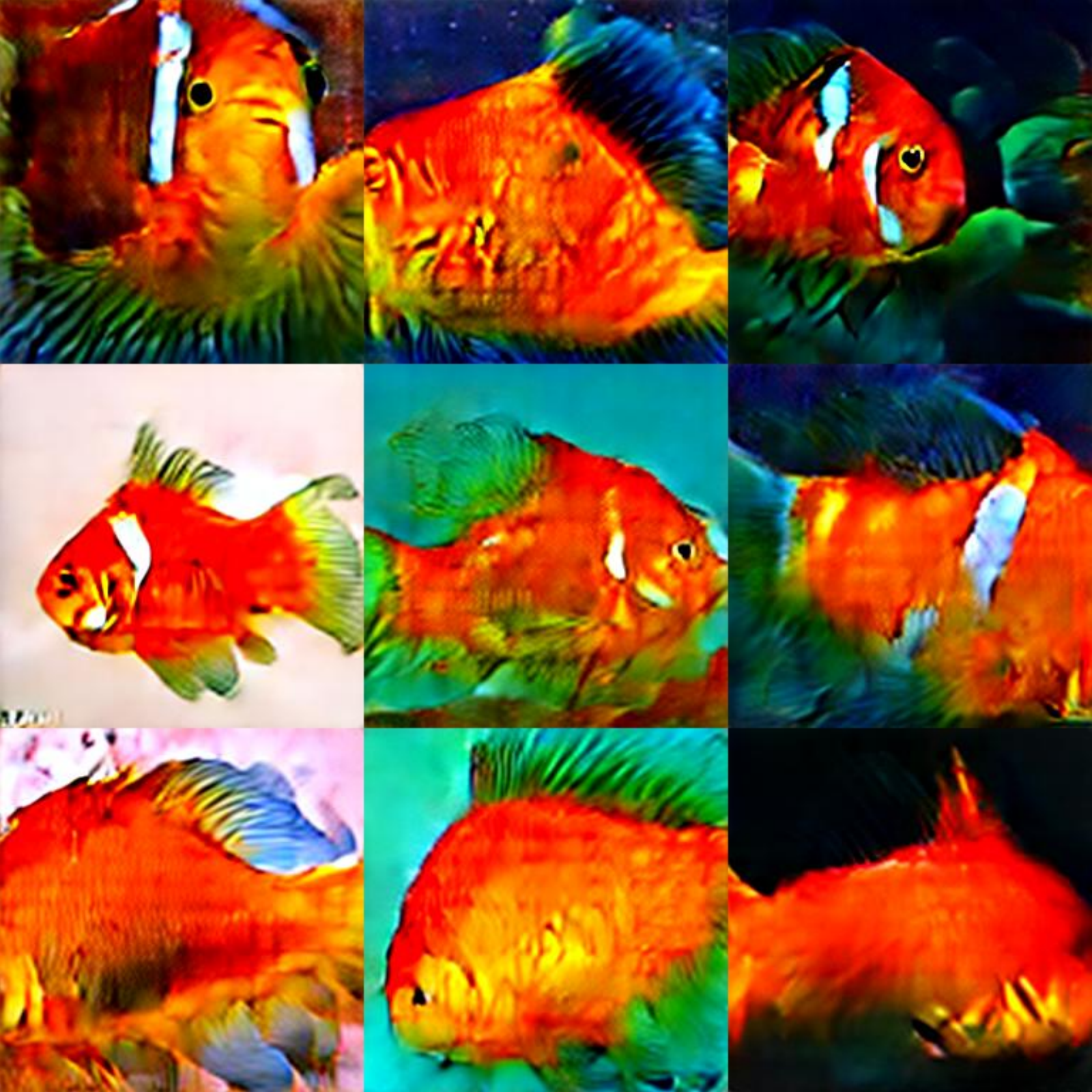}}  &
        \adjustbox{valign=m}{\includegraphics[width=0.28\linewidth]{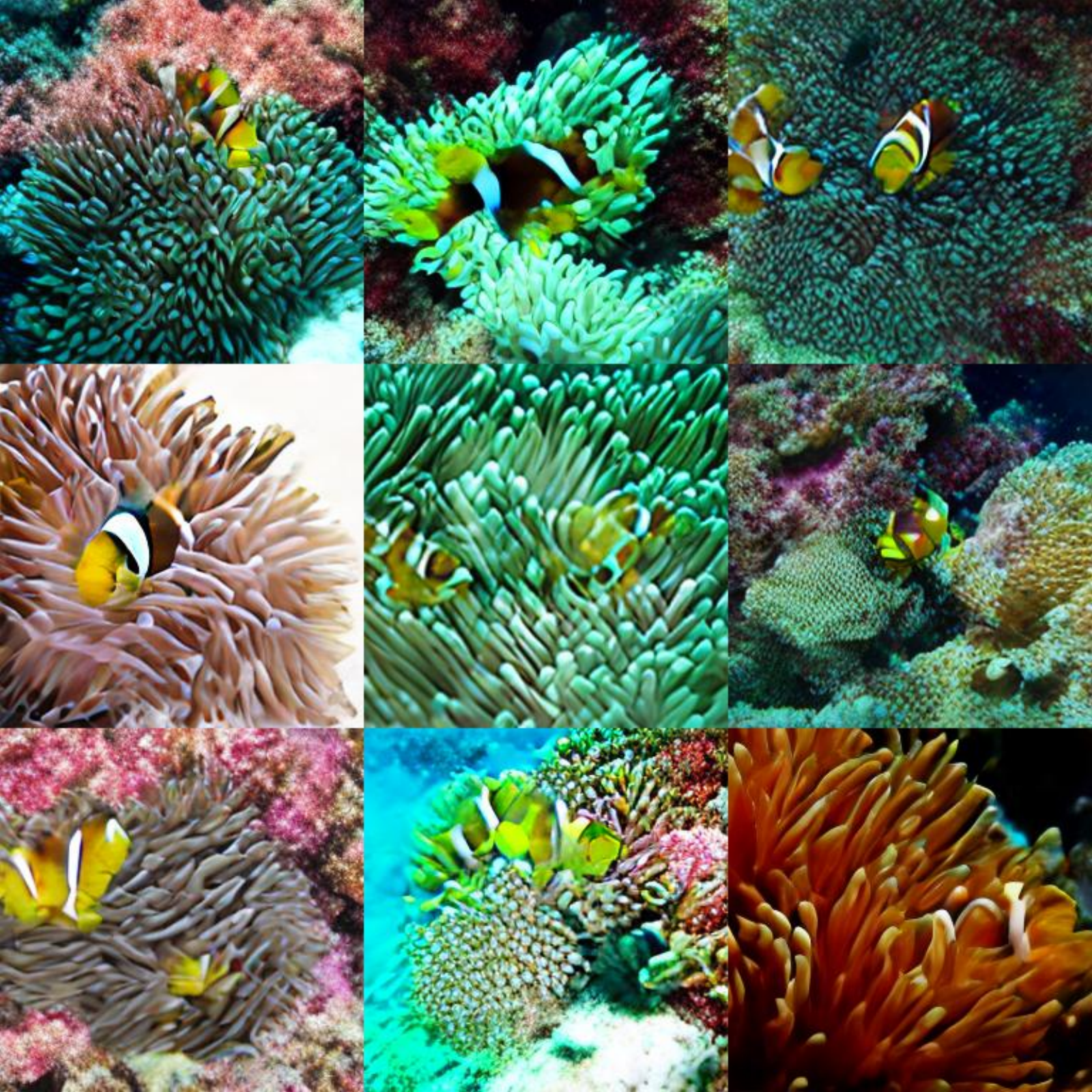}} \\
  \end{tabular}
  \caption{\textbf{Unlearning performance on ImageNet-256 (Goldfish).} Samples for the \textit{Forget} and \textit{Retain} classes. GA suffers from severe structural corruption (u-FID: 79.89) to achieve concept suppression. In contrast, UOT-Unlearn achieves robust erasure (85.08\% PUL) while preserving generative fidelity (u-FID: 20.16) relative to the baseline (FID: 11.57). u-FID is computed over 36 aquatic classes.}
  \label{fig:qualitative_comparison}
  \vspace{-8pt}
\end{figure}

\paragraph{Experimental Results.}
Our framework demonstrates a strong ability to navigate the trade-off between concept suppression and generative fidelity. As summarized in \cref{tab:main_results} and \cref{fig:quantitative_results}, our method achieves consistently high PUL scores across all targeted classes while maintaining an exceptionally low u-FID. This performance indicates that the cost-driven probability redistribution effectively suppresses the forget class without degrading the learned data distribution. Qualitative results in \cref{fig:qualitative_results} further validate that our approach eliminates targeted semantics while preserving the structural integrity of remaining concepts.

High-resolution generation tasks in a class-conditional setting highlight the limitations of standard unlearning objectives. As reported in \cref{fig:qualitative_comparison}, achieving even moderate concept erasure with GA on ImageNet-256 targeting the `Goldfish' concept fails to preserve the structural fidelity of the retain distribution, inflating the u-FID to 79.89.
To ensure a structured probability redistribution, we restrict the conditioning labels during the generator update to a localized subset of 37 aquatic classes, comprising the forget target and 36 semantically adjacent concepts. This constraint encourages the model to remap the penalized concepts into relevant retain domains, facilitating a smooth semantic shift rather than introducing unintended distributional shifts. Consequently, our method yields a robust PUL of 85.08\% while successfully limiting u-FID degradation to 20.16, generating diverse, high-fidelity scenes even when strictly conditioned on the forget class.

\subsection{Ablation Study} \label{sec:exp_ablation}

\begin{figure}[t]
  \centering
  \begin{subfigure}{0.48\linewidth}
    \centering
    \includegraphics[trim=0cm 0cm 0cm 0.75cm, clip, width=\linewidth]{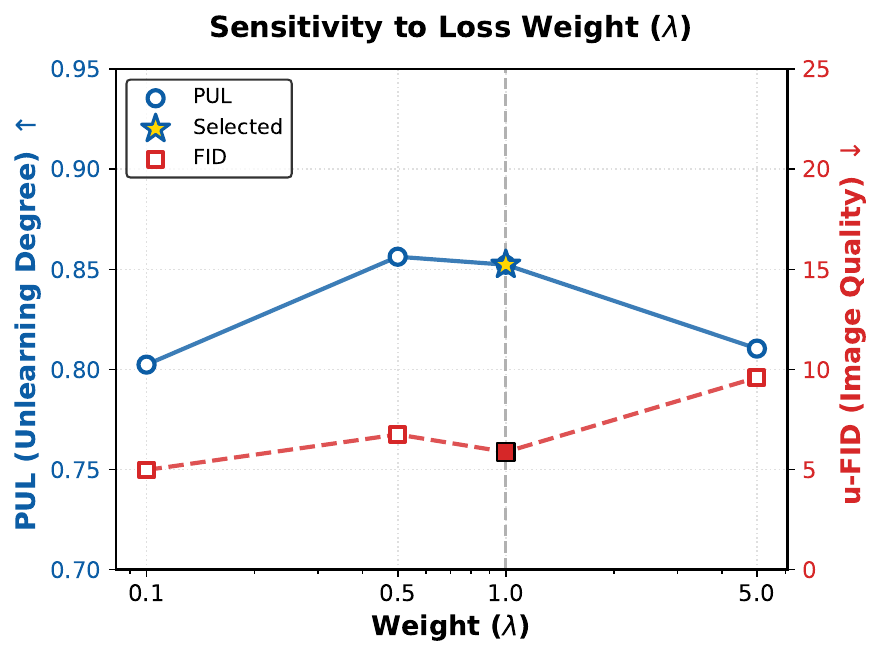}
    \caption{Sensitivity to Loss Weight ($\lambda$)}
    \label{fig:ablation_lambda}
  \end{subfigure}
  \hfill
  \begin{subfigure}{0.48\linewidth}
    \centering
    \includegraphics[trim=0cm 0cm 0cm 0.75cm, clip, width=\linewidth]{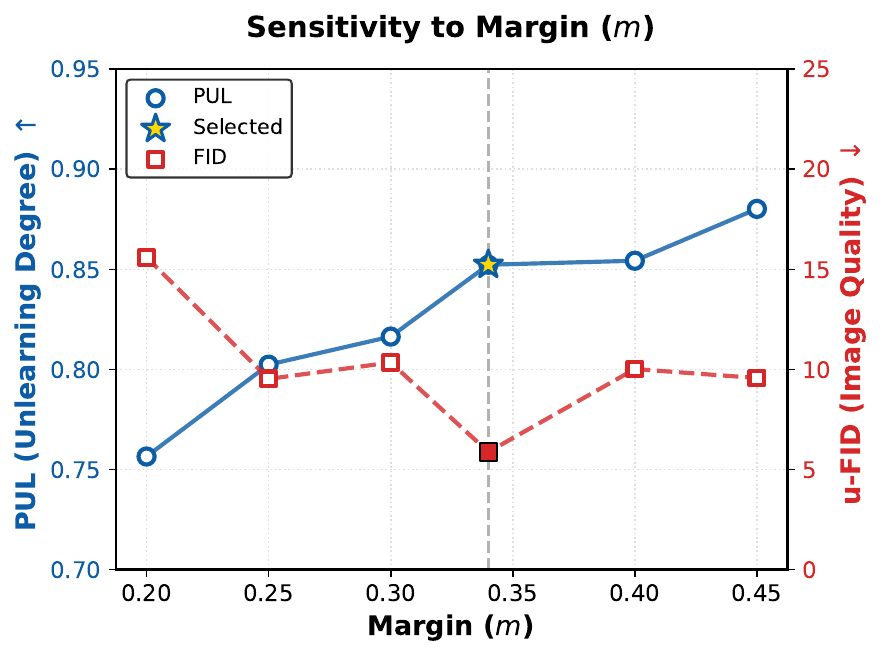}
    \caption{Sensitivity to Margin ($m$)}
    \label{fig:ablation_margin}
  \end{subfigure}
  
  \caption{\textbf{Ablation study of key hyperparameters} for our UOT framework, conducted on CTM model for CIFAR-10 with target class 8. We independently vary the forget loss weight $\lambda$ (left) and the semantic distance margin $m$ (right). The star marker ($\star$) denotes our optimal configuration, which effectively balances the concept erasure efficacy (PUL $\uparrow$) and the generative fidelity (u-FID $\downarrow$).}
  \label{fig:ablation_study}
  \vspace{-8pt}
\end{figure}

We investigate the sensitivity of the proposed UOT framework to its core hyperparameters: the \textit{forget loss weight} $\lambda$ and the \textit{feature distance margin} $m$ (\cref{eq:unlearning_cost}), by targeting Class 8 on the CTM architecture. The parameter $\lambda$ controls the strength of the unlearning term. As illustrated in \cref{fig:ablation_lambda}, setting $\lambda = 1.0$ provides the most favorable trade-off between concept erasure (PUL) and generative fidelity (u-FID). Relaxing this penalty ($\lambda = 0.1$) provides insufficient optimization signal for target removal. Conversely, an excessively large weight ($\lambda = 5.0$) destabilizes the unlearning process. Rather than monotonically improving erasure, this over-penalization alters the learned flow mapping, leading to a simultaneous degradation in both PUL and u-FID.

The margin $m$ defines the feature-space boundary required to accurately isolate and displace the target distribution. We observe an empirical optimum at $m = 0.34$, where the target concepts are successfully erased without corrupting adjacent semantic regions (\cref{fig:ablation_margin}). A conservative margin ($m \le 0.30$) fails to fully encapsulate the forget set. Consequently, the model generates residual target semantics that mismatch the true retain distribution, increasing the u-FID. On the other hand, an overly aggressive margin ($m \ge 0.40$) forces the unlearning objective to encroach upon neighboring retain classes. This excessive margin degrades the structural fidelity of non-target domains, demonstrating that a precisely calibrated margin is essential for localized concept removal.

\section{Conclusion}

We introduced a machine unlearning framework for one-step generative models. 
Rather than adapting iterative denoising-based strategies, we cast concept erasure as an unbalanced optimal transport (UOT) problem and incorporate a transport cost directly into the single-pass mapping. This formulation redistributes probability mass away from the forget region without requiring access to real retain data.

Empirical results show that baselines adapted to the one-step regime consistently exhibit an efficacy–fidelity trade-off. Strong suppression often leads to significant deviations from the original data distribution, whereas milder interventions leave residual target semantics. In contrast, our method consistently attains high unlearning performance (PUL) while maintaining limited deviation in u-FID. These observations indicate that the UOT regularization induces a structured redistribution of probability mass rather than large-scale distortion of the learned distribution.

Because the proposed objective operates on the synthetic probability flow without architectural modifications, it is compatible with different classes of fast generative models. Viewing unlearning as constrained probability transport offers a clear formulation
for balancing removal strength and distributional consistency. Future work includes exploring structured probability redistribution across related semantic classes within highly structured latent spaces, such as large-scale hierarchical datasets, as well as analyzing the theoretical stability of the unlearning process under alternative cost formulations.

%% file: text_sup.tex
\section{Implementation Details}
\label{sec:implementation_details}

\subsection{Experimental Settings}
In our unlearning experiments, we evaluate our approach on both unconditional and class-conditional generation tasks across different resolutions. To ensure reproducibility, all generative models, classifiers, and feature extractors used in our experiments are adopted from publicly available pretrained checkpoints.

\begin{itemize}
    \item \textbf{CIFAR-10 \cite{krizhevsky2009learning} (Unconditional Generation):} We utilize Consistency Trajectory Model (CTM) \cite{kim2023consistency} and Meanflow \cite{geng2025mean} models as unconditional generators. We focus on the class unlearning task—erasing a specific single class from the dataset—and conduct separate unlearning experiments targeting Class 1, Class 6, and Class 8.
    
    \item \textbf{ImageNet-256 \cite{russakovsky2015imagenet} (High-Resolution Class-Conditional Setting):} We employ the Meanflow \cite{geng2025mean} model. Rather than unlearning across all 1,000 classes, we restrict the conditioning labels during the generator update to a localized subset of 37 aquatic classes. Specifically, we choose ``goldfish'' as our target forget class, while the remaining 36 classes serve as semantically adjacent concepts. This localized constraint is intentionally designed to encourage the model to remap the forgotten concept into relevant domains, facilitating a smooth semantic shift rather than introducing unintended distributional shifts.
    
    \item \textbf{Forget Anchor ($\mu_f$) Calculation:} To compute the unlearning cost $c_{\mathrm{ul}}$ and construct the forget anchor (centroid) $\mu_f$ in the feature space, we employ a pretrained feature extractor $f(\cdot)$ on a small subset of sampled images from the forget class. Specifically, we utilize 512 images for the CIFAR-10 dataset and 260 images for the ImageNet-256 dataset (representing exactly 20\% of the approximately 1,300 available training images per class). This demonstrates that our proposed method requires minimal data to accurately identify and erase the target concept.
    
    \item \textbf{Strict Unlearning Scenario:} Furthermore, we operate under a strict unlearning scenario where the original retain data is completely inaccessible, and only the fully pretrained checkpoint is available. This specific checkpoint constraint necessitates structural modifications to the baseline methods, which are detailed in Section \ref{sec:baseline_methods}.
\end{itemize}

\subsection{Pretrained Models and Evaluation Networks}

\paragraph{Pretrained Generative Models} 
For unconditional generation on the CIFAR-10 dataset, we adopt the pretrained CTM (FID: 1.73) and Meanflow (FID: 2.80) checkpoints. For high-resolution conditional generation on ImageNet-256, we utilize the pretrained conditional Meanflow checkpoint (SiT-XL/2) with an initial FID of 3.43. Following the ECCV anonymity policy, exact links to the pretrained models are omitted and will be added later.

\paragraph{Evaluation Networks} 
To rigorously evaluate the efficacy of our unlearning framework, we employ the following pretrained evaluation networks:
\begin{itemize}
    \item \textbf{Classifier (for PUL calculation):} For CIFAR-10, we use a pretrained DenseNet-121 model which achieves an accuracy of 94.06\%. For ImageNet-256, we employ a pretrained ViT-L/16 model, achieving an accuracy of 88.55\%.
    \item \textbf{FID Calculation:} We compute the Fréchet Inception Distance (FID) and unlearned-FID (u-FID) using the standard Inception-v3 network.
\end{itemize}

\subsection{UOT-Unlearn Implementation}

\paragraph{Network Architecture} 
Our proposed UOT-Unlearn framework consists of a generator and a discriminator. Since the primary objective is to unlearn specific concepts from a pretrained model, the generator is directly initialized with the weights of the fully pretrained CTM or Meanflow checkpoint. For the discriminator, we adopt the exact architecture proposed in the standard Unbalanced Optimal Transport (UOT) generative model \cite{choi2023generative}. Specifically, we use their small ResNet-based discriminator variant, featuring a base channel size of 64, LeakyReLU (0.2) activations, and a minibatch standard deviation layer.

\paragraph{Feature Extractor (for Centroid Calculation)} 
To extract feature representations and compute the forget centroid, we utilize the penultimate layer of a pretrained ResNet-56 model (94.37\% accuracy) for CIFAR-10, and a pretrained ResNet-50 model (81.19\% accuracy) for ImageNet-256. To ensure a strictly fair and unbiased evaluation, we intentionally employ these ResNet-based feature extractors during the unlearning phase, which are distinctly separated from the classifiers (DenseNet-121 and ViT-L/16) used later for calculating the Probability of Unlearning (PUL) metric.

\paragraph{Training Hyperparameters} 
For the CIFAR-10 unlearning experiments, we use a batch size of 128. Regarding the exponential moving average (EMA) of the generator weights, we disable EMA during the CTM unlearning process, whereas we set the EMA decay rate to 0.99 for the Meanflow unlearning process. For the high-resolution ImageNet-256 experiments, we utilize a batch size of 8 and set the EMA decay rate to 0.999. All other optimization settings strictly follow the default configuration provided in the original UOT implementation (e.g., Adam optimizer with learning rates of $1.6 \times 10^{-4}$ for the generator and $1.0 \times 10^{-4}$ for the discriminator).

\section{Baseline Methods}
\label{sec:baseline_methods}

To evaluate the effectiveness of our proposed framework, we compare it against several representative unlearning baselines. For fair comparison under our strict unlearning scenario (where only a single pretrained checkpoint is available and retain data is inaccessible), we establish the following baseline setups.

\subsection{Gradient Ascent (GA)}
For the Gradient Ascent baseline, we directly maximize the original training loss on the entire training set of the target class, denoted as $\mathcal{D}_{\text{forget}}$. Since we apply this to the Consistency Trajectory Model (CTM) and Meanflow models, the GA loss is simply the negative of their respective training objectives:
\begin{equation}
    \mathcal{L}_{GA}(\theta) = - \mathbb{E}_{x \sim \mathcal{D}_{\text{forget}}} [\mathcal{L}_{\text{train}}(\theta; x)]
\end{equation}
where $\mathcal{L}_{\text{train}}$ denotes either the CTM or Meanflow loss. 

For the class-conditional setting (i.e., ImageNet-256), the training objective inherently depends on both the image data and the class condition. Let $y_f$ denote the target forget class label. The GA loss is naturally extended to maximize the expected conditional training loss over the forget data:
\begin{equation}
    \mathcal{L}_{GA}(\theta) = - \mathbb{E}_{x \sim \mathcal{D}_{\text{forget}}} [\mathcal{L}_{\text{train}}(\theta; x, y_f)]
\end{equation}
where $\mathcal{L}_{\text{train}}(\theta; x, y_f)$ represents the conditional training objective for a given image-label pair. This formulation ensures the model parameters are updated to diverge from the original conditional distribution associated with $y_f$.

\subsection{Variational Diffusion Unlearning (VDU) \cite{panda2024variational}}
The Variational Diffusion Unlearning (VDU) framework formulates the unlearning objective by integrating the negative training loss with a parameter penalty term to prevent catastrophic degradation. The general VDU loss is defined as:
\begin{equation}
    \mathcal{L}_{VDU}(\theta) = -(1-\gamma) \mathcal{L}_{\text{train}}(\theta; \mathcal{D}_{\text{forget}}) + \gamma \sum_{i=1}^{d} \frac{(\theta_i - \mu_i^*)^2}{2\sigma_i^{*2}}
\end{equation}
where $\gamma$ is the penalty weight, $d$ is the total number of model parameters, and $\mu_i^*$ and $\sigma_i^*$ represent the empirical mean and standard deviation of the $i$-th parameter, respectively. For our experiments, we set $\gamma = 0.005$ for CTM and $\gamma = 0.001$ for Meanflow.

To compute the parameter statistics ($\mu_i^*$ and $\sigma_i^*$), the original VDU method relies on multiple historical checkpoints saved during the initial pretraining phase. However, under our strict setting where only a single fully pretrained checkpoint is provided, we adopt a practical alternative strategy to estimate these statistics. Starting from the pretrained checkpoint, we fine-tune the model on the full training dataset for 4 epochs (with a learning rate of $1 \times 10^{-6}$ and a batch size of 128). We save a checkpoint at the end of every epoch. Combined with the initial pretrained checkpoint, this yields a total of 5 checkpoints (from 0 to 4 epochs), which we use to compute the required empirical mean and variance for the penalty term.

\subsection{Selective Amnesia (SA) \cite{heng2023selective}}
\label{sec:sa_baseline}
Existing machine unlearning methods, such as Selective Amnesia (SA) and Saliency Unlearning (SalUn), are primarily designed for conditional generative models. To apply these methods to our unconditional generators, we first construct a common base unlearning objective ($\mathcal{L}_{\text{base}}$) that utilizes noise mapping and a pseudo-retain dataset.

\paragraph{Common Base Objective ($\mathcal{L}_{\text{base}}$)}
Since our unconditional generator lacks explicit class conditions to locate the target concept, we use a pretrained auxiliary classifier to filter the prior noise space. Let $\mathcal{Z}_{\text{forget}}$ denote the set of prior noise vectors $x_0 \sim \mathcal{N}(0, I)$ that the generator maps to images classified as the target concept. We introduce a \textit{Noise Mapping Loss} to force the generator to output pure random noise $\epsilon$ for these specific latents:
\begin{equation}
\mathcal{L}_{\text{forget\_noise}}(\theta) = \mathbb{E}_{x_0 \sim \mathcal{Z}_{\text{forget}}, \epsilon \sim \mathcal{N}(0,I)} \left[ \left\| G_\theta(x_0) - \epsilon \right\|_2^2 \right]
\end{equation}
Simultaneously, we formulate a retain loss using a fixed pseudo-retain dataset, $\mathcal{D}_{\text{pr}}$. We construct this set offline by generating images with the original generator and filtering out the target concept using the auxiliary classifier. We apply the standard training objective to these samples:
\begin{equation}
\mathcal{L}_{\text{retain}}(\theta) = \mathbb{E}_{x \sim \mathcal{D}_{\text{pr}}} \left[ \mathcal{L}_{\text{train}}(\theta; x) \right]
\end{equation}
The base unlearning objective is the weighted sum of these two terms: $\mathcal{L}_{\text{base}} = \alpha \mathcal{L}_{\text{forget\_noise}} + \beta \mathcal{L}_{\text{retain}}$. We empirically tuned the hyperparameters via grid search, utilizing $\alpha \in \{0.05, 0.1\}$ and $\beta \in \{0.5, 5.0\}$.

\paragraph{SA Objective}
To constrain parameter deviation during unlearning, SA incorporates an Elastic Weight Consolidation (EWC) penalty into the base loss. This requires the computation of the Fisher Information Matrix (FIM), denoted as $F$. We compute the FIM using the entire generated pseudo-dataset, $\mathcal{D}_{\text{pseudo}} = \mathcal{D}_{\text{pr}} \cup \mathcal{D}_{\text{pf}}$. In practice, a diagonal approximation is adopted where each diagonal element $F_i$ corresponding to parameter $\theta_i$ is computed as $F_i = \mathbb{E}_{x \sim \mathcal{D}_{\text{pseudo}}} [ ( \nabla_{\theta_i} \mathcal{L}_{\text{train}}(\theta_{\text{pre}}; x) )^2 ]$. The final SA objective is:
\begin{equation}
\mathcal{L}_{\text{SA}}(\theta) = \mathcal{L}_{\text{base}} + \frac{\lambda_{\text{SA}}}{2} \sum_i F_i (\theta_i - \theta_{\text{pre}, i})^2
\end{equation}
where $\theta_{\text{pre}}$ represents the pretrained weights, and the penalty scale $\lambda_{\text{SA}}$ is empirically set (e.g., $5.0$ or $1000.0$ based on the model scale).

\subsection{Saliency Unlearning (SalUn) \cite{fan2024salun}}
SalUn prevents catastrophic forgetting by restricting weight updates to only the most salient parameters associated with the target concept. Rather than modifying the loss function directly with a penalty, SalUn optimizes the same base objective ($\mathcal{L}_{\text{base}}$) defined in Section \ref{sec:sa_baseline}, but applies a binary gradient mask $\mathbf{M}$ during the update step:
\begin{equation}
\mathbf{M} = \mathbb{I} \left( \left| \left. \nabla_{\boldsymbol{\theta}} \mathbb{E}_{\mathbf{x} \sim \mathcal{D}_{\text{forget}}} \left[ \mathcal{L}_{\text{train}}(\boldsymbol{\theta}; \mathbf{x}) \right] \right|_{\boldsymbol{\theta}=\boldsymbol{\theta}_{pre}} \right| \ge \gamma \right)
\end{equation}

While the original SalUn paper recommends a sparsity ratio of 50\% for multi-step diffusion models, we empirically found that unfreezing such a large proportion of weights severely disrupts the single-pass mapping of one-step models, leading to a collapse in generation quality. Thus, we strictly adjust the threshold $\gamma$ to enforce a sparsity ratio of 95\%, selecting only the top 5\% of the parameters to be updated. During optimization, we freeze the non-salient weights and update only the salient parameters selected by the mask:
\begin{equation}
\boldsymbol{\theta}_{t+1} = \boldsymbol{\theta}_t - \eta \left( \mathbf{M} \odot \nabla_{\boldsymbol{\theta}} \mathcal{L}_{\text{base}} \right)
\end{equation}
where $\odot$ is the element-wise product and $\eta$ is the learning rate.

\section{Visualization}

\subsection{CIFAR-10}

\begin{figure*}[tb]
    \centering
    \setlength{\tabcolsep}{2pt}
    \begin{tabular}{cc}
        
        \begin{minipage}[c]{0.04\linewidth}
            \centering
            \rotatebox{90}{\small Pretrained}
        \end{minipage} & 
        \begin{minipage}[c]{0.6\linewidth}
            \includegraphics[width=\linewidth]{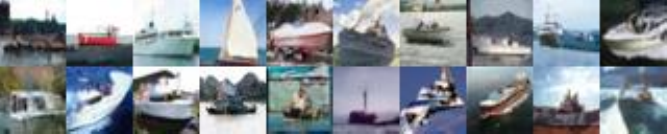}
        \end{minipage} \\
        \noalign{\vspace{0.3cm}}
        
        \begin{minipage}[c]{0.04\linewidth}
            \centering
            \vspace{0.5cm}
            \rotatebox{90}{\small GA} \\
            \vspace{0.8cm}
            \rotatebox{90}{\small VDU} \\
            \vspace{0.8cm}
            \rotatebox{90}{\small SA} \\
            \vspace{0.8cm}
            \rotatebox{90}{\small SalUn} \\
            \vspace{0.3cm}
        \end{minipage} & 
        \begin{minipage}[c]{0.6\linewidth}
            \includegraphics[width=\linewidth]{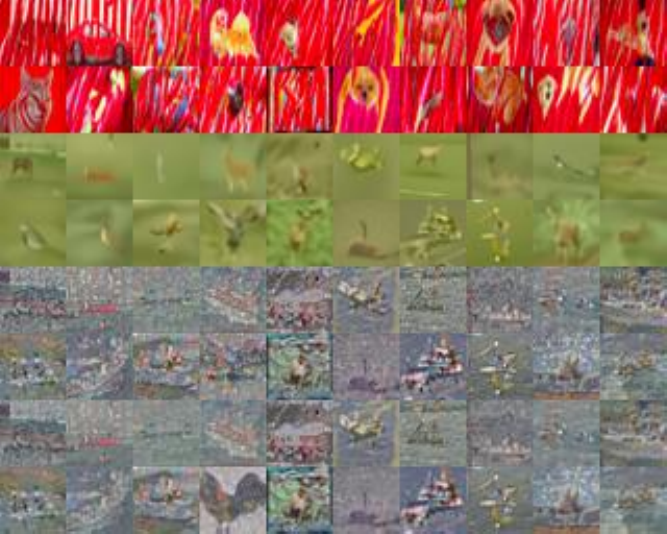}
        \end{minipage} \\

        \begin{minipage}[c]{0.04\linewidth}
            \centering
            \rotatebox{90}{\small \textbf{UOT-Unlearn}}
        \end{minipage} & 
        \begin{minipage}[c]{0.6\linewidth}
            \includegraphics[width=\linewidth]{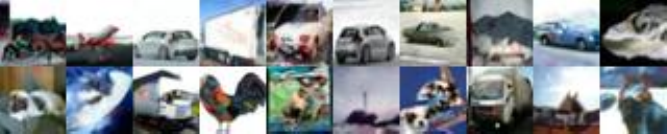}
        \end{minipage} \\
        
    \end{tabular}
    \caption{Qualitative comparison of unlearning methods on the Ship class (Class 8) using the Consistency Trajectory Model (CTM). Generated under fixed seed setting, this figure illustrates how the target class is erased across different approaches. Notably, unlike other baselines generate corrupted images, UOT-Unlearn demonstrates a distinct tendency to transition the forgotten concept into features of other classes while preserving the overall spatial layout.}
    \label{fig:supp_qualitative_class8_ctm}
\end{figure*}

\begin{figure*}[tb]
    \centering
    \begin{minipage}{0.48\linewidth}
        \centering
        \textbf{Pretrained} \\[1mm]
        \includegraphics[width=\linewidth]{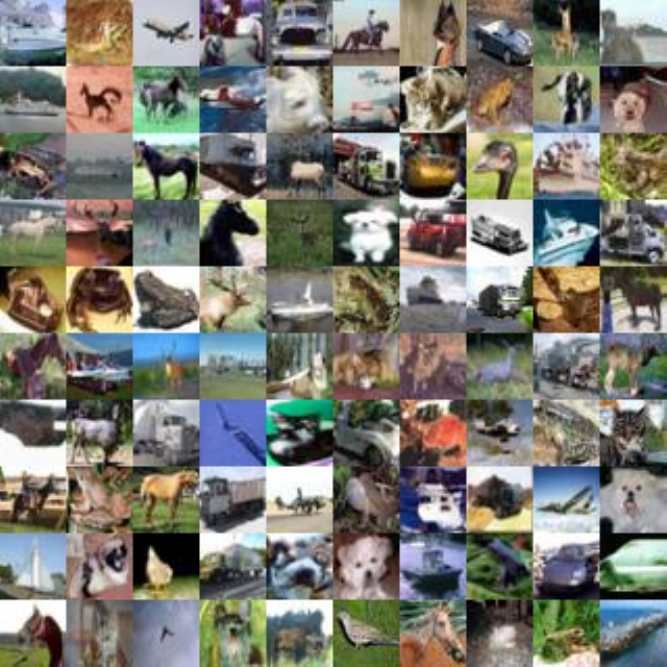}
    \end{minipage}
    \hfill
    \begin{minipage}{0.48\linewidth}
        \centering
        \textbf{UOT-Unlearn (CTM)} \\[1mm]
        \includegraphics[width=\linewidth]{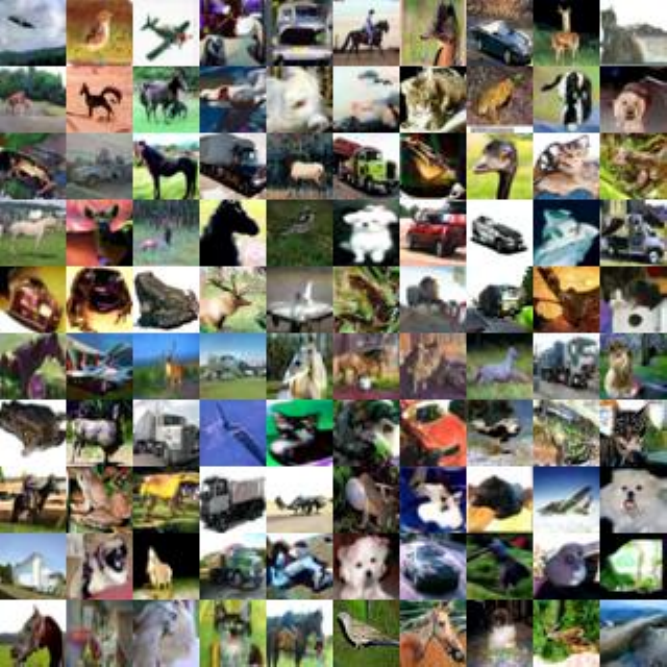}
    \end{minipage}
    \vspace{-2mm}
    \caption{Visual comparison of images generated from the pretrained CTM versus our UOT-Unlearn-applied CTM trained to forget the Ship class (Class 8). The results demonstrate that UOT-Unlearn successfully erases the forget class, seamlessly transitioning it into features of other classes, while strictly preserving the visual quality and structural layout of the remaining retain classes.}
    \label{fig:supp_random_10x10_ctm}
\end{figure*}

\begin{figure*}[tb]
    \centering
    \setlength{\tabcolsep}{2pt}
    \begin{tabular}{cc}
        
        \begin{minipage}[c]{0.04\linewidth}
            \centering
            \rotatebox{90}{\small Pretrained}
        \end{minipage} & 
        \begin{minipage}[c]{0.6\linewidth}
            \includegraphics[width=\linewidth]{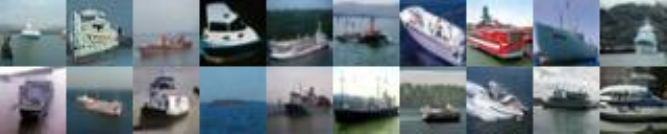}
        \end{minipage} \\
        \noalign{\vspace{0.3cm}}
        
        \begin{minipage}[c]{0.04\linewidth}
            \centering
            \vspace{0.5cm}
            \rotatebox{90}{\small GA} \\
            \vspace{0.8cm}
            \rotatebox{90}{\small VDU} \\
            \vspace{0.8cm}
            \rotatebox{90}{\small SA} \\
            \vspace{0.8cm}
            \rotatebox{90}{\small SalUn} \\
            \vspace{0.3cm}
        \end{minipage} & 
        \begin{minipage}[c]{0.6\linewidth}
            \includegraphics[width=\linewidth]{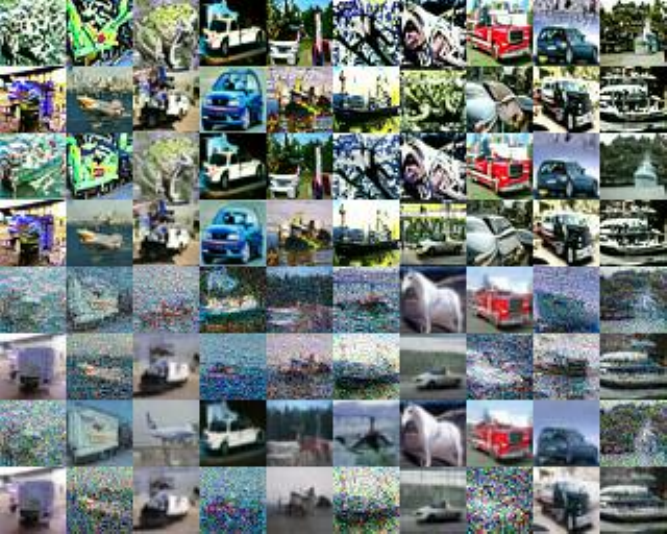}
        \end{minipage} \\

        \begin{minipage}[c]{0.04\linewidth}
            \centering
            \rotatebox{90}{\small \textbf{UOT-Unlearn}}
        \end{minipage} & 
        \begin{minipage}[c]{0.6\linewidth}
            \includegraphics[width=\linewidth]{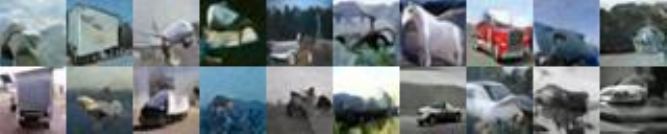}
        \end{minipage} \\
        
    \end{tabular}
    \caption{Qualitative comparison of unlearning methods on the Ship class (Class 8) using the Meanflow (MF). Generated under fixed seed setting, this figure illustrates how the target class is erased across different approaches.}
    \label{fig:supp_qualitative_class8_mf}
\end{figure*}

\begin{figure*}[tb]
    \centering
    \begin{minipage}{0.48\linewidth}
        \centering
        \textbf{Pretrained} \\[1mm]
        \includegraphics[width=\linewidth]{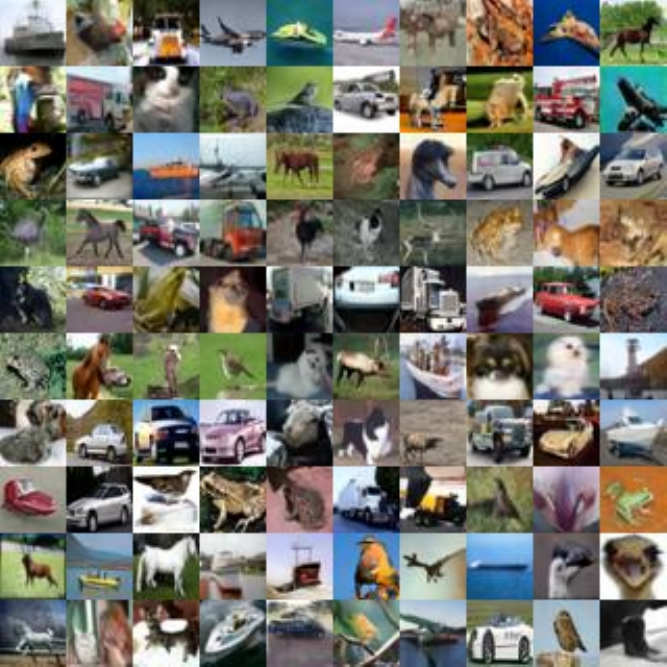}
    \end{minipage}
    \hfill
    \begin{minipage}{0.48\linewidth}
        \centering
        \textbf{UOT-Unlearn (Meanflow)} \\[1mm]
        \includegraphics[width=\linewidth]{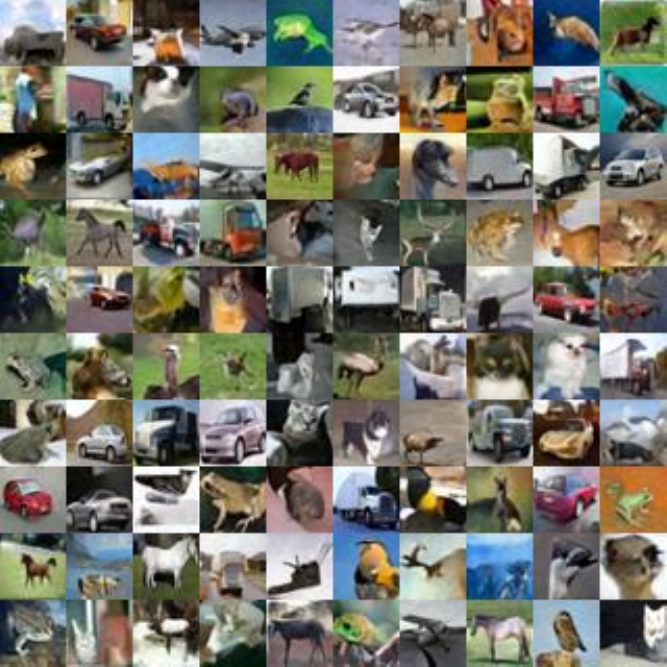}
    \end{minipage}
    \vspace{-2mm}
    \caption{Visual comparison of randomly generated images from the Pretrained model and our UOT-Unlearn (MF) trained to forget the Ship class (Class 8), using the same initial noises.}
    \label{fig:supp_random_10x10_mf}
\end{figure*}

We present visual results on the CIFAR-10 dataset based on the configurations that achieved the best performance in Table 2 of the main text. \cref{fig:supp_qualitative_class8_ctm} and \cref{fig:supp_qualitative_class8_mf} compare the unlearning patterns of the forget class across the baselines and our UOT-Unlearn, using the CTM and MF models, respectively. Furthermore, \cref{fig:supp_random_10x10_ctm} and \cref{fig:supp_random_10x10_mf} display how randomly generated images from the pretrained generators change after applying UOT-Unlearn. Overall, these visualizations demonstrate that our method not only erases the target concept but also naturally shifts its generation toward the retained classes, successfully preserving the structural layout of the original scenes.

\subsection{ImageNet 256$\times$256}

\begin{figure*}[tb]
    \centering
    \begin{subfigure}{\linewidth}
        \centering
        \includegraphics[width=0.53\linewidth]{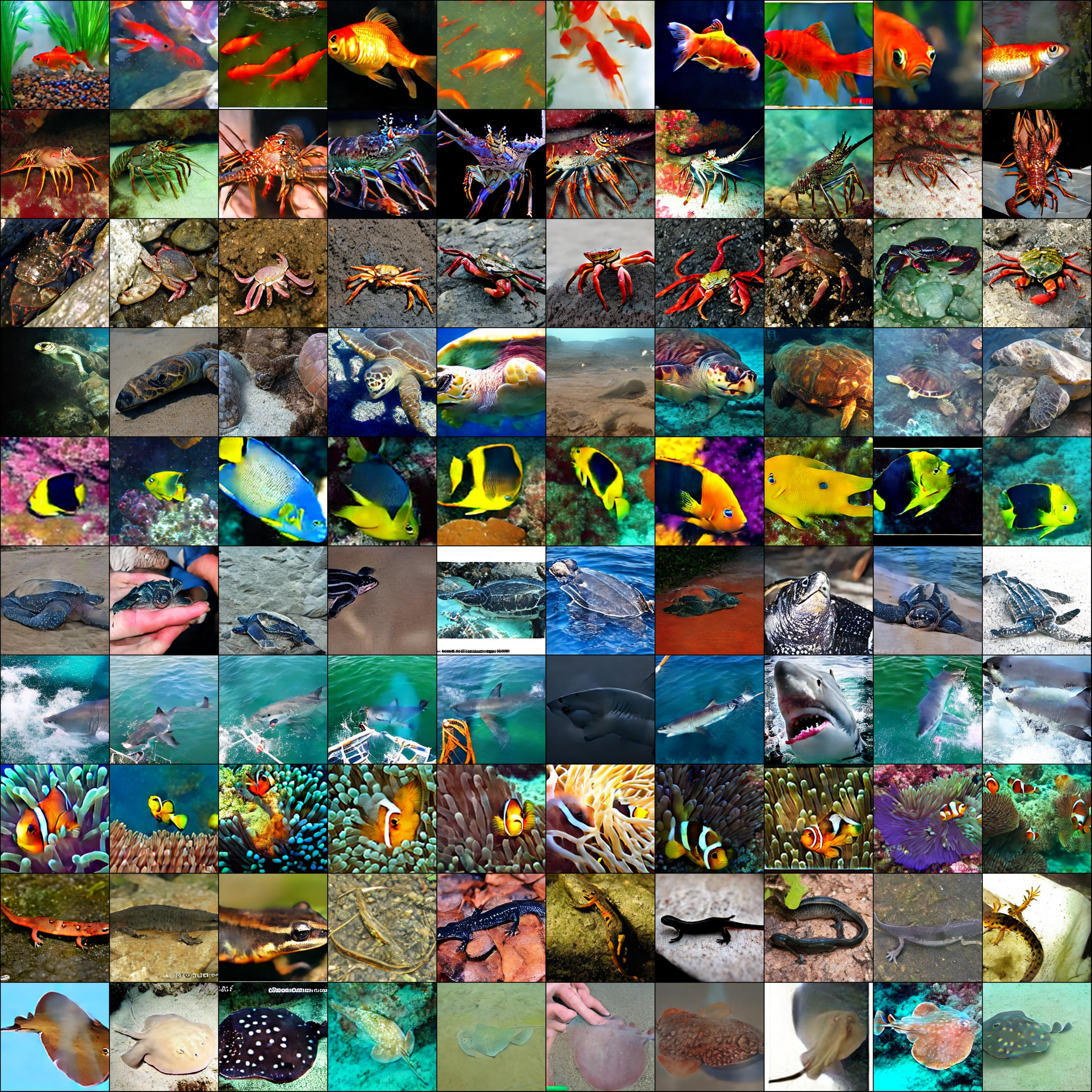}
        \caption{\textbf{Pretrained}}
        \label{fig:supp_imagenet_pre}
    \end{subfigure}
    \vspace{0.4cm}
    \begin{subfigure}{\linewidth}
        \centering
        \includegraphics[width=0.53\linewidth]{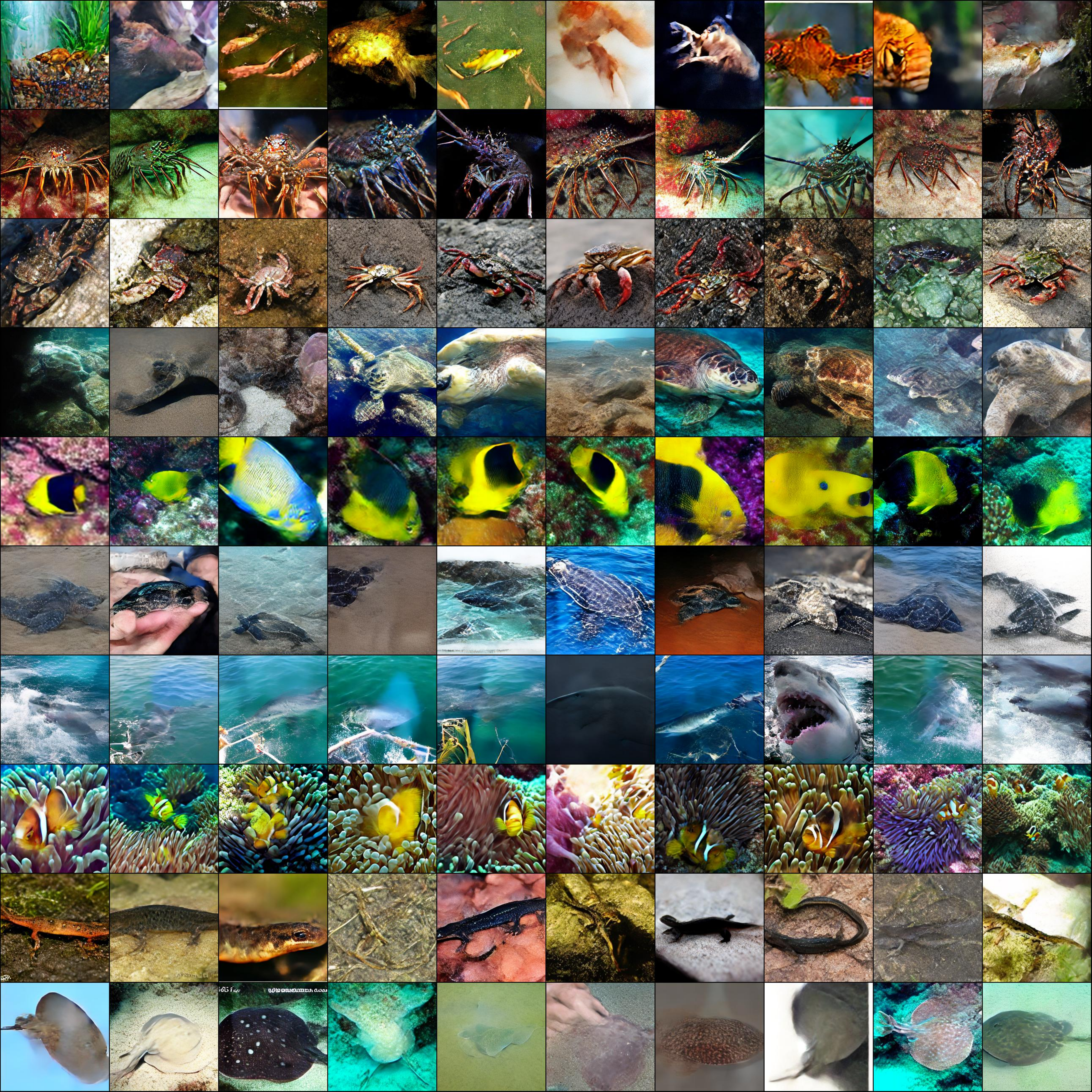}
        \caption{\textbf{UOT-Unlearn (Meanflow)}}
        \label{fig:supp_imagenet_uot}
    \end{subfigure}
    \vspace{-0.8cm}
    \caption{Unlearning results on the ImageNet 256$\times$256 dataset using the MeanFlow (MF) model. The top row of each generated batch displays the Goldfish class (forget class), while the subsequent bottom rows show 9 other aquatic animal classes (retain classes) using the exact same initial noises.}
    \label{fig:supp_imagenet_10x10}
\end{figure*}

As an extension to the results shown in Fig. 4 of the main text, we provide additional visualization results on the ImageNet $256\times256$ dataset using the Meanflow model (\cref{fig:supp_imagenet_10x10}). Using identical initial noises, we compare the pretrained model and UOT-Unlearn across the target forget class (goldfish, top row) and 9 aquatic retain classes. Visually, UOT-Unlearn effectively erases the specific features of the forget class while largely preserving the structural layouts of the retain classes. These qualitative observations align with our robust metrics (85.08\% PUL; u-FID 20.16 vs. pretrained FID 11.57).


\clearpage

\clearpage

\clearpage

\clearpage

\clearpage

\clearpage